\pdfoutput=1
\documentclass[11pt]{article}

\usepackage{style/acl_arr}

\usepackage{times}
\usepackage{latexsym}

\usepackage[T1]{fontenc}

\usepackage[utf8]{inputenc}

\usepackage{microtype}

\usepackage{algorithm}
\usepackage[noend]{algpseudocode}
\usepackage{varwidth}
\usepackage{multirow}
\usepackage{tabularx}
\usepackage{subcaption}
\usepackage{amssymb}
\usepackage{amsmath}
\usepackage{amsthm}
\usepackage{amsfonts}
\usepackage{bbm}
\usepackage{bm}
\usepackage{booktabs}
\usepackage{graphicx}
\usepackage{multirow}
\usepackage{multicol}
\usepackage{enumitem}
\usepackage{mdwlist}
\usepackage{csquotes}

\setlist{nosep}

\title{Adapting Coreference Resolution Models through Active Learning}
\author{Michelle Yuan$^\dagger$\ \ \ \ Patrick Xia$^\ddagger$\ \ \ \ Chandler May$^\ddagger$\\
{\bf Benjamin Van Durme}$^\ddagger$\ \ \ \ {\bf Jordan Boyd-Graber}$^\dagger$\\
    Human Language Technology Center of Excellence\\
    $^\dagger$University of Maryland\ \ \ \ \ $^\ddagger$Johns Hopkins University\\
    \texttt{myuan@cs.umd.edu paxia@cs.jhu.edu jbg@umiacs.umd.edu} \\}

\date{}

\newif\ifcomment\commenttrue
\usepackage[a-1b]{pdfx}

\usepackage{framed}
\usepackage{mdwlist}
\usepackage{siunitx}
\usepackage{latexsym}
\usepackage{colortbl}
\usepackage{xcolor}
\usepackage{nicefrac}
\usepackage{booktabs}
\usepackage{fnpct}
\usepackage{amsfonts}
\usepackage[T1]{fontenc}
\usepackage{bold-extra}
\usepackage{amsmath}
\usepackage{amssymb}
\usepackage{bm}
\usepackage{graphicx}
\usepackage{mathtools}
\usepackage{microtype}
\usepackage{multirow}
\usepackage{multicol}
\usepackage{xpatch}
\usepackage{latexsym,comment}
\usepackage[normalem]{ulem}

\newcommand*{\missingreference}{{\Huge \colorbox{red}{?reference?}}}
\newcommand*{\missingcitation}{{\Huge \colorbox{red}{?citation?}}}

\makeatletter
\xpatchcmd{\@setref}{\bfseries}{\missingreference}{}{}
\def\@citex[#1]#2{\leavevmode
    \let\@citea\@empty
    \@cite{\@for\@citeb:=#2\do
        {\@citea\def\@citea{,\penalty\@m\ }%
            \edef\@citeb{\expandafter\@firstofone\@citeb\@empty}%
            \if@filesw\immediate\write\@auxout{\string\citation{\@citeb}}\fi
            \@ifundefined{b@\@citeb}{\hbox{\reset@font\missingcitation}%
                \G@refundefinedtrue
                \@latex@warning
                {Citation `\@citeb' on page \thepage \space undefined}}%
            {\@cite@ofmt{\csname b@\@citeb\endcsname}}}}{#1}}
\makeatother

\newcommand{\gem}[1]{\mbox{\textsc{gem}}}
\newcommand{\abr}[1]{\textsc{#1}}

\DeclareMathOperator*{\argmax}{arg\,max}

\newcommand{\ex}[1]{\mbox{exp}\left\{ #1\right\} }

\newcommand{\hidetext}[1]{}
\newcommand{\ignore}[1]{}

\ifcomment
    \newcommand{\pinaforecomment}[3]{\colorbox{#1}{\parbox{.8\linewidth}{#2: #3}}}

    \newcommand{\prtodo}[1]{\pinaforecomment{lightblue}{pr}{#1}}
    \newcommand{\prtodoi}[1]{\pinaforecomment{lightblue}{pr}{#1}}
\else
    \newcommand{\pinaforecomment}[3]{}
    \newcommand{\prtodo}[1]{}
    \newcommand{\prtodoi}[1]{}
\fi

\newcommand{\smallurl}[1]{ \begin{tiny}\url{#1}\end{tiny}}

\definecolor{lightblue}{HTML}{3cc7ea}
\definecolor{CUgold}{HTML}{CFB87C}
\definecolor{grey}{rgb}{0.95,0.95,0.95}
\definecolor{ceil}{rgb}{0.57, 0.63, 0.81}
\definecolor{UMDred}{HTML}{ed1c24}
\definecolor{UMDyellow}{HTML}{ffc20e}

\newcommand{\fone}{$F_1$}

\usepackage[normalem]{ulem}

\newcommand{\avgfone}{\abr{Avg}~$\textsc{F}_1$}
\newcommand{\coref}{\abr{cr}}
\newcommand{\icoref}{\abr{ic{\small oref}}}
\newcommand{\spanbert}{\abr{s{\small pan}bert}}
\newcommand{\spanbertlarge}{\spanbert-\textsc{\small large}-\textsc{\small cased}}
\newcommand{\ctof}{\abr{c{\small 2}f-coref}}
\newcommand{\ontonotes}{\abr{o{\small nto}n{\small otes}}}
\newcommand{\preco}{\abr{p{\small re}c{\small o}}}
\newcommand{\qbcoref}{\abr{qbc{\small oref}}}

\newcommand{\given}{\, | \,}

\newcommand{\ent}[1]{\text{H}_{\text{#1}}}
\newcommand{\entity}[2]{[#1]\textsubscript{#2}}

\graphicspath{ {figures/}{auto_commit_fig/}{auto_fig/} }

\begin{document}
\maketitle
\begin{abstract}
Neural coreference resolution models trained on one dataset may not
transfer to new, low-resource domains.
Active learning mitigates this problem by sampling a small subset of data for annotators to label.
While active learning is well-defined for classification tasks, its application
to coreference resolution is neither well-defined nor fully understood.
This paper explores how to actively label coreference, examining sources of model uncertainty and document reading costs.
We compare uncertainty sampling strategies and their advantages through thorough error analysis.
In both synthetic and human experiments, labeling
spans within the same document is more effective than annotating spans across documents.
The findings contribute to a more realistic development of coreference resolution models.

\end{abstract}
\section{Introduction}
\label{sec:intro}

Linguistic expressions are coreferent if they refer to the same
entity.
The computational task of discovering coreferent mentions is coreference
resolution (\coref).
Neural models~\citep{lee-2018,joshi-2020} are \abr{sota}
on \ontonotes~{\small 5.0}~\citep{pradhan-2013} but
cannot immediately generalize to other datasets.
Generalization is difficult because domains differ in content, writing style, and annotation
guidelines.
To overcome these challenges, models need copiously labeled,
in-domain data~\citep{bamman-2020}.

\begin{figure*}
    \center
    \includegraphics[width=\linewidth]{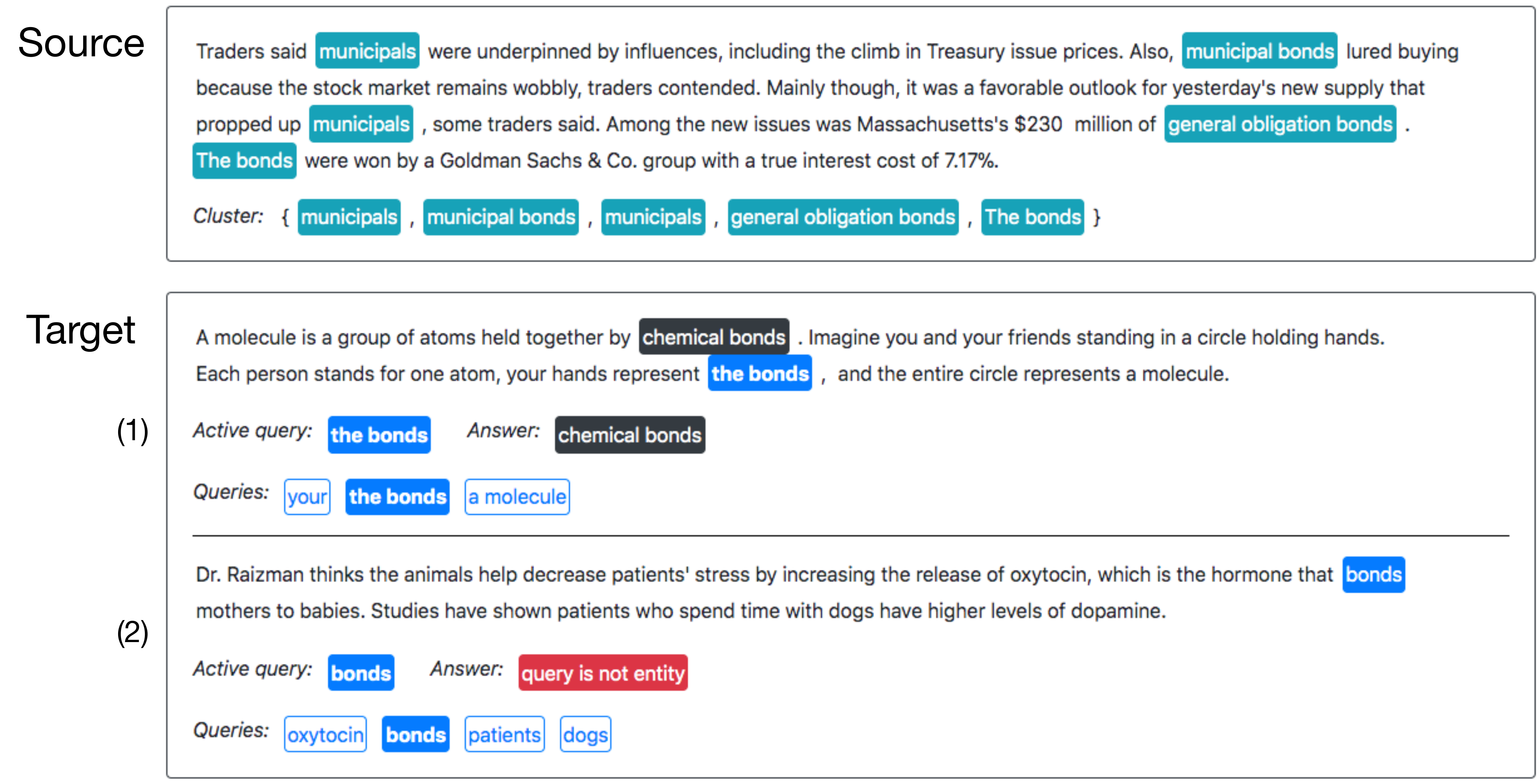}
    \caption{
        \coref{} models are trained on \textbf{source} domain \ontonotes{}, which contains data like
        news articles. The \textbf{source}
        document links ``the bonds'' to ``municipal bonds''.
        In a \textbf{target} domain like \preco{}~\citep{chen-2018-preco}, ``the bonds'' may no longer
        have the same meaning. It can refer to
        ``chemical bonds'' (Document 1) or not be considered an entity (Document 2).
        A solution is to continue training the \textbf{source} model on more spans
        from the \textbf{target} domain. Active learning helps select
        ambiguous spans, like ``the bonds'', for the user to label on this interface (Section~\ref{ssec:human_labeling}).
    }
    \label{fig:example}
  \end{figure*}

Despite expensive labeling costs, adapting \coref{} is crucial for applications like
uncovering information about proteins in biomedicine~\citep{kim-2012}
and distinguishing entities in legal documents~\citep{gupta-2018}.
Ideally, we would like to quickly and cheaply adapt the model without repeatedly relying on an excessive amount of annotations to retrain the model.
To reduce labeling cost, we investigate active learning~\citep{settles-2009}
for \coref{}.  Active learning aims to reduce annotation costs by
intelligently selecting examples to label.
Prior approaches use active learning to improve the model within the same domain~\citep{gasperin-2009,sachan-2015} without considering adapting to new data distributions.
For domain adaptation in \coref{},
\citet{zhao-2014} motivate the use of active learning
to select out-of-distribution examples.
A word like ``the bonds'' refers to municipal bonds in
\ontonotes{} but links to ``chemical bonds'' in another
domain (Figure~\ref{fig:example}).
If users annotate the antecedents of ``the
bonds'' and other ambiguous entity mentions, then these labels
help adapt a model trained on \ontonotes{} to new domains.

Active learning for \coref{} adaptation is well-motivated, but the implementation is neither straightforward nor well-studied.
First, \coref{} is a span detection and clustering task,
so selecting which spans to label is more complicated than
choosing independent examples for text classification.
Second, \coref{} labeling involves closely reading the
documents.
Labeling more spans within the same context is more efficient.  However, labeling more spans across different documents increases data diversity and may
improve model transfer.
How should we balance these competing objectives?

Our paper extends prior work in active
learning for \coref{} to the problem of coreference model
transfer~\citep{xia-2021}:
\begin{enumerate*}
    \item We generalize the \emph{clustered entropy} sampling
        strategy~\citep{li-2020} to include uncertainty in mention detection. We analyze the effect of
        each strategy on coreference model transfer.
    \item We investigate the trade-off between labeling and reading
    through simulations and a real-time user study.
    Limiting annotations to the same document
    increases labeling throughput and decreases volatility in model
    training.
\end{enumerate*}
Taken together, these contributions offer a blueprint for faster creation of \coref{} models
across domains.\footnote{\url{https://github.com/forest-snow/incremental-coref}}

\section{Problem: Adapting Coreference}
\label{sec:model}

\citet{lee-2018} introduce \ctof{}, a neural model that outperforms prior rule-based systems.
It assigns an antecedent $y$ to mention span $x$. The set
$\mathcal{Y}(x)$ of
possible antecedent spans include a dummy antecedent $\epsilon$ and all spans
preceding $x$.
If span~$x$ has
no antecedent, then $x$ should be assigned to $\epsilon$.  Given entity mention $x$,
the model learns a distribution over its candidate antecedents in $\mathcal{Y}(x)$,
\begin{equation}
    P(Y=y) = \frac{\ex{s(x,y)}}{\sum_{y' \in \mathcal{Y}(x)} \ex{s(x,y')}}.
    \label{eq:antecedent}
\end{equation}
The scores $s(x,y)$ are computed by the model's pairwise scorer
(Appendix~\ref{ssec:neural}).

\coref{} models like \ctof{} are typically trained on \ontonotes{}.
Recent work in \coref{} improves upon \ctof{} and has \abr{sota} results on \ontonotes{}~\citep{wu-2020-corefqa,joshi-2020}.
However, annotation guidelines and the underlying text differ across domains. As a result, these \coref{} models cannot
immediately transfer to other datasets.
For different domains, spans could hold
different meanings or link to different entities.
\citet{xia-2021} show the benefits of \emph{continued training} where a model
trained on \ontonotes{} is further trained on the target dataset.
For several target domains, continued training from \ontonotes{} is stronger than
training the model from scratch, especially when the training dataset is small.

Their experiments use an incremental variant of \ctof{} called
\icoref{}~\citep{xia-2020}.
While \ctof{} requires $\Theta(n)$ memory to simultaneously access all spans in
the document and infer a span's antecedent,
\icoref{} only needs constant memory to predict a span's entity
cluster.
Despite using less space, \icoref{} retains the same accuracy as \ctof{}.
Rather than assigning $x$ to antecedent $y$, \icoref{}
 assigns $x$ to cluster $c$ where $c$ is from a set of observed entity clusters $\mathcal{C}$,
\begin{equation}
    P(C=c) = \frac{\ex{s(x,c)}}{\sum_{c' \in \mathcal{C}} \ex{s(x,c')}}.
    \label{eq:cluster}
\end{equation}
As the algorithm processes spans in the
document, each span is either placed in a cluster from $\mathcal{C}$ or
added to a new cluster.
To learn the distribution over clusters (Equation~\ref{eq:cluster}), the algorithm first creates a cluster
representation that is an aggregate of span
representations over spans that currently exist in the cluster.
With cluster and span representations, individual spans and entity clusters are
mapped into a shared space. Then, we can compute $s(x,c)$ using the same pairwise scorer as before.

\citet{xia-2021} show that continued training is useful for domain adaptation
but assume that labeled data already exist in the target
domain.
However, model transfer is more critical when annotations are scarce.
Thus, the question becomes: how can we adapt \coref{} models without requiring a
large, labeled dataset?
Our paper investigates active learning as a potential solution.  Through active learning, we reduce labeling costs by sampling and annotating a small subset of ambiguous spans.

\section{Method: Active Learning}
\label{sec:active}
Neural models achieve high accuracy for \ontonotes{} but cannot quickly
adapt to new datasets because of shifts in domain or
annotation standards~\citep{poot-2020}.
To transfer to new domains, models need substantial in-domain, labeled data.
In low-resource situations, \coref{} is infeasible for real-time
applications.
To reduce the labeling burden, active learning may
target spans that most confuse the model.
Active learning for domain adaptation~\citep{rai-2010} typically proceeds as follows:
begin with a model
trained on source data,
sample and label $k$ spans from
documents in the
target domain based on a strategy, and
train the model
on labeled data.

This labeling setup may appear straightforward to apply to \coref{}, but there are some tricky details.
The first complication is that---unlike text classification---\coref{} is a \emph{clustering} task.
Early approaches in active learning for \coref{} use \emph{pairwise
annotations}~\citep{miller-2012,sachan-2015}.
Pairs of spans are sampled and the annotator labels
whether each pair is coreferent.
The downside to pairwise annotations is that it requires many labels.
To label the antecedent
of entity mention $x$, $x$ must be compared to every
candidate span in the document. \citet{li-2020} propose a new scheme
called \emph{discrete annotations}.
Instead of sampling pairs of spans, the active learning strategy samples
individual spans.
Then, the annotator only has to find and label first antecedent of $x$ in the
document, which bypasses the multiple pairwise comparisons. Thus, we use
discrete annotations to minimize labeling.

To further improve active learning for \coref{}, we consider the following
issues.
First, the \coref{} model has different scores for mention detection and linking,
but prior active learning methods only considers linking. Second,
labeling \coref{} requires time to read the document context.
Therefore, we explore important aspects of active learning
 for adapting \coref{}: model uncertainty
(Section~\ref{ssec:uncertainty}), and the balance between reading and labeling
(Section~\ref{ssec:tradeoff}).

\subsection{Uncertainty Sampling}
\label{ssec:uncertainty}
A well-known active learning strategy is uncertainty sampling.
A common measure of uncertainty is the entropy in the distribution of the model's
predictions for a given example~\citep{lewis-1994}.
Labeling uncertain examples improves
accuracy for tasks like text classification~\citep{settles-2009}.
For \coref{}, models have multiple components, and computing uncertainty is not
as
straightforward.
Is uncertainty over where mentions are located more important than linking
spans?  Or the other way around?
Thus, we investigate different sources of \coref{}
model uncertainty.

\subsubsection{Clustered Entropy}
\label{ssec:clust-ent}

To sample spans for learning \coref{}, \citet{li-2020} propose a strategy called
\emph{clustered entropy}.  This metric scores the uncertainty in the entity
cluster assignment of a mention span $x$. If $x$ has \emph{high} clustered entropy,
then it should be labeled to help the model learn its antecedents. Computing clustered entropy
requires the probability that $x$ is assigned to an entity cluster.
\citet{li-2020} use \ctof{}, which only gives probability of $x$ being
assigned to antecedent $y$. So, they define $P(C=c)$ as the sum of antecedent probabilities $P(Y=y)$,
\begin{equation}
    P(C=c) = \sum_{y \in C \cap \mathcal{Y}(x)} P(Y=y).
\end{equation}
Then, they define clustered entropy as,
\begin{equation}
    \ent{}(x) = -\sum_{c \in \mathcal{C}} P(C=c) \log P(C=c).
    \label{eq:li-clust-ent}
\end{equation}
The computation of clustered entropy in Equation~\ref{eq:li-clust-ent} poses two issues.
First, summing the probabilities may not accurately represent the model's probability of linking $x$ to $c$.
There are other ways to aggregate the probabilities (e.g.~taking the maximum).
\ctof{} never computes cluster probabilities to make predictions, so it is not obvious how $P(C=c)$ should be computed for clustered entropy.
Second, Equation~\ref{eq:li-clust-ent} does
not consider mention detection.  For \ontonotes{}, this is not an issue
because singletons (clusters of size 1) are not annotated and mention detection
score is implicitly included in $P(Y=y)$.
For other datasets containing singletons, the model should disambiguate
singleton clusters from non-mention spans.

To resolve these issues,
we make the following changes. First, we use \icoref{} to obtain cluster
probabilities. \icoref{} is a
mention clustering model so it already has probabilities over
entity clusters~(Equation~\ref{eq:cluster}).  Second, we explore other forms of
maximum entropy sampling.  Neural \coref{} models have scorers for mention
detection and clustering.  Both scores should be considered
to sample spans that confuse the model.
Thus, we propose more strategies to target uncertainty in mention detection.

\subsubsection{Generalizing Entropy in Coreference}
\label{ssec:entropy}

To generalize entropy sampling, we first formalize mention
detection and clustering. Given span $x$, assume $X$ is the random
variable encoding whether
$x$ is an entity mention (1) or not (0).
In Section~\ref{sec:model}, we assume that the cluster distribution
$P(C)$ is independent
of $X$: $P(C) = P(C \given X)$.\footnote{A side effect of \ontonotes{}
  models lacking singletons.} In other words, Equation~\ref{eq:cluster} is actually computing $P(C=c \given X=1)$.
We sample top-$k$ spans with the following strategies.

\paragraph{ment-ent} Highest mention detection entropy:
\begin{align}
    \ent{MENT}(x)
        &= \ent{}{(X)} \\
        &= - \sum_{i=0}^1 P(X=i) \log P(X=i). \nonumber
\end{align}
The probability $P(X)$ is
computed from normalized mention scores $s_m$ (Equation~\ref{eq:unary}).
\textbf{Ment-ent} may sample spans that challenge mention detection (e.g.
class-ambiguous words like ``park'').
The annotator can clarify whether spans are entity mentions to improve mention
detection.

\paragraph{clust-ent} Highest mention clustering entropy:
\begin{align}
    \ent{CLUST}(x)
        &= \ent{}(C \given X = 1) \\
        &= -\sum_{c \in \mathcal{C}} P(C = c \given X =
    1) \log \nonumber \\
    & \qquad P(C = c \given X = 1). \nonumber
\end{align}
\textbf{Clust-ent} looks at clustering scores
without explicitly addressing mention detection.
Like in \ontonotes{}, all spans are assumed to be entity mentions.
The likelihood $P(C =c \given X = 1)$ is given by \icoref{}
(Equation~\ref{eq:cluster}).

\paragraph{cond-ent} Highest conditional entropy:
\begin{equation}
\begin{split}
    \ent{COND}(x)
        &= \ent{}(C \given X) \\
        &= \sum_{i=0}^1 P(X=i) \ent{}(C \given X = i) \\
        &= P(X=1) \ent{}(C \given X=1) \\
        &= P(X=1) \ent{CLUST}(x).
\end{split}
\end{equation}
We reach the last equation because there is no uncertainty in
clustering $x$ if $x$ is not an entity mention and $\ent{}(C \given X=0) = 0$.
\textbf{Cond-ent} takes the uncertainty of mention detection into
account.  So, we may sample more pronouns because they are
obviously mentions but difficult to cluster.

\paragraph{joint-ent} Highest joint entropy:
\begin{align}
    \ent{JOINT}(x)
        &= \ent{}(X,C) = \ent{}(X) + \ent{}(C \given X) \nonumber \\
        &= \ent{MENT}(x) + \ent{COND}(x).
\end{align}
\textbf{Joint-ent} may sample spans that are difficult to detect as entity mentions
\emph{and} too confusing to
cluster.
This sampling strategy most closely aligns with the uncertainty of the
training objective. It may also fix any imbalance between mention detection and
linking~\citep{wu-2020}.

\subsection{Trade-off between Reading and Labeling}
\label{ssec:tradeoff}

For \coref{}, the annotator reads the document context
to label the antecedent of a mention span.
Annotating and reading spans from different documents may slow down labeling,
but restricting sampling to the same
document may cause redundant labeling~\citep{miller-2012}.
To better understand this trade-off, we explore different
configurations with~$k$, the number of annotated spans, and~$m$, the maximum number of
documents being read.  Given source model $h_0$ already
fine-tuned on \ontonotes{}, we adapt $h_0$ to a target
domain through active learning (Algorithm~\ref{alg:active}):

\paragraph{Scoring}
To sample~$k$ spans from unlabeled
data~$\mathcal{U}$ of the target domain, we score spans with an
active learning strategy~$S$.
Assume $S$
scores each span through an \textit{acquisition model}~\citep{lowell-2019}.
For the acquisition model, we use $h_{t-1}$, the model fine-tuned from the last cycle.
The acquisition score quantifies the span's importance given
$S$ and the acquisition model.

\paragraph{Reading}
Typically, active learning samples~$k$ spans with
the highest acquisition scores.
To constrain~$m$, the number of documents read, we
find the documents of the~$m$ spans with highest
acquisition scores and only sample spans from those documents.
Then, the~$k$ sampled spans will belong to at most~$m$ documents.
If $m$ is set to ``unconstrained'', then we simply sample the $k$
highest-scoring spans,
irrespective of the document boundaries.

Our approach resembles \citet{miller-2012} where they sample spans based on
highest uncertainty and continue sampling from the same document until
uncertainty falls below a threshold.
Then, they sample the most uncertain span
from a new document. We modify their method because the uncertainty
threshold will vary for different
datasets and models.  Instead, we use the number of documents read to
control context switching.

\paragraph{Labeling}
An oracle (e.g.,
human annotator or gold data) labels the antecedents of sampled spans with discrete
annotations (Section~\ref{sec:active}).

\paragraph{Continued Training}
We combine data labeled from current and past cycles. We
train the source model $h_0$ (which is already trained on \ontonotes{}) on the labeled target data. We do not continue training a model from a past active learning cycle because
it may be biased from only training on scarce target
data~\citep{ash-2020-warmstart}.

\begin{algorithm}[t]
\caption{Active Learning for Coreference}
\begin{algorithmic}[1]
    \Require Source model $h_0$, Unlabeled data $\mathcal{U}$,
    Active learning strategy $S$,
    No.~of cycles $T$,
    No.~of labeled spans $k$, Max. no. of read docs $m$
    \State Labeled data $\mathcal{L}=\{\}$
    \For {cycles $t=1, \dots, T$}
        \State $a_x \gets$ Score span $x \in \mathcal{U}$ by $S(h_{t-1}, x)$
        \State $\mathcal{Q} \gets$ Sort ($\downarrow$) $x \in \mathcal{U}$ by scores $a_x$
        \State $\mathcal{Q}_m \gets$ Top-$m$ spans in $\mathcal{Q}$
        \State $\mathcal{D} \gets \{\bm{d}_x \given x \in \mathcal{Q}_m \}$ where
        $\bm{d}_x$ is doc of $x$
        \State $\widetilde{\mathcal{Q}} \gets$ Filter $\mathcal{Q}$ s.t.~spans belong to
        $\bm{d} \in \mathcal{D}$
        \State $\widetilde{\mathcal{Q}}_k \gets$ Top-$k$ spans in
        $\widetilde{\mathcal{Q}}$
        \State $\mathcal{L}_k \gets$ Label antecedents for
        $\widetilde{\mathcal{Q}}_k$
        \State $\mathcal{L} \gets \mathcal{L} \cup \mathcal{L}_k$
        \State $h_t \gets$ Continue train $h_0$ on $\mathcal{L}$
    \EndFor
    \Return $h_T$
\end{algorithmic}
\label{alg:active}
\end{algorithm}

\section{Active Learning for \coref{} through Simulations and Humans}
\label{sec:experiments}

We run experiments to understand two important factors of active learning for
\coref{}: sources of model
uncertainty (Section~\ref{ssec:uncertainty}) and balancing reading against
labeling
(Sections~\ref{ssec:tradeoff}).
First, we simulate active learning on \preco{} to compare sampling strategies
based on various forms of uncertainty
(Section~\ref{ssec:sampling_results}).
Then, we set up a user study to investigate how humans perform when labeling
spans from fewer or more documents from \preco{} (Section~\ref{ssec:human_labeling}).
Specifically, we analyze their annotation time and throughput.
Finally, we run large-scale simulations on \preco{} and
\qbcoref{} (Section~\ref{ssec:adaptation}).
We explore different combinations of sampling strategies and labeling
configurations.

\paragraph{Models}
In all experiments, the source model is the best checkpoint of \icoref{} model
 trained on
\ontonotes{}~\citep{xia-2020}
with
\spanbertlarge~\citep{joshi-2020} encoder.
For continued training on the target dataset, we optimize with a fixed parameter
configuration (Appendix~\ref{ssec:params}).
We evaluate models on
\avgfone{}, the averaged \fone{} scores of \abr{muc}~\citep{vilain-1995}, $\abr{b}^3$~\citep{bagga-1998},
and $\abr{ceaf}_{\phi4}$~\citep{luo-2005}.
For all synthetic experiments, we simulate active learning with gold data
substituting as an annotator. However, gold mention boundaries are not used when
sampling data. The model scores spans that are likely to be entity
mentions for inference, so we limit the active learning candidates to this pool of high-scoring
spans.
For each active learning simulation, we repeat five runs with different random seed initializations.
\paragraph{Baselines}
We compare the proposed sampling strategies (Section~\ref{ssec:entropy})
along with \textbf{li-clust-ent}, which is
clustered entropy from \citet{li-2020} (Equation~\ref{eq:li-clust-ent}).
Active learning is frustratingly less effective than random sampling in many settings~\citep{lowell-2019}, so we include two random baselines in our
simulation. \textbf{Random} samples from all spans in the
documents. \textbf{Random-ment}, as well as other strategies, samples only from the
pool of likely (high-scoring) spans.
Thus, \textbf{random-ment} should be a stronger
baseline than \textbf{random}.

\begin{figure}[!t]
    \centering
    \includegraphics[width=\linewidth]{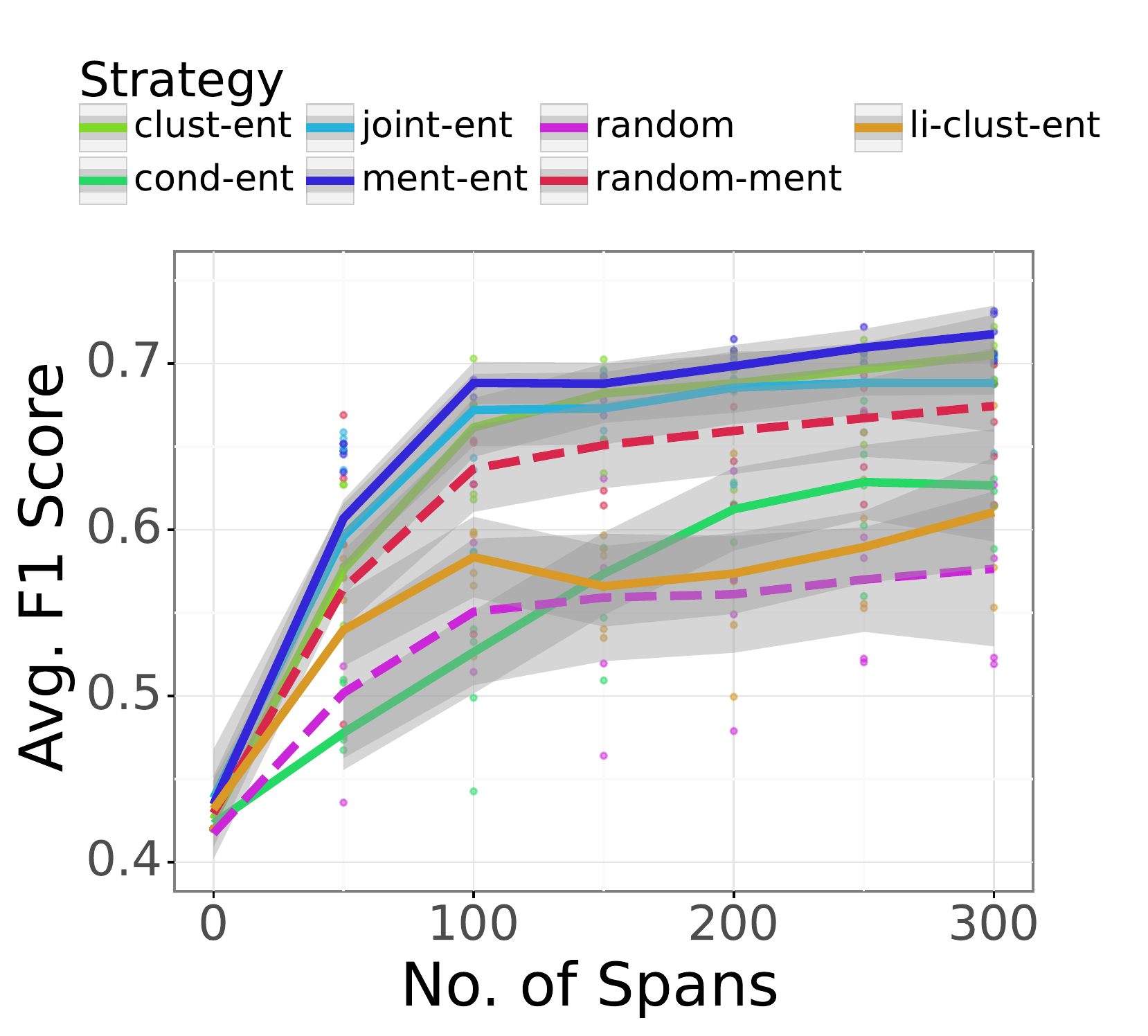}
    \caption{Test \avgfone{} on \preco{} for each strategy.
    On each
    cycle,
    fifty spans from one document are sampled and labeled.
    We repeat each
    simulation five times.
    \textbf{Ment-ent},
    \textbf{clust-ent}, and \textbf{joint-ent} are most  effective
    while
    \textbf{random} hurts the model the most.
    }
    \label{fig:preco_simple}
\end{figure}

\begin{figure}[!t]
    \centering
    \includegraphics[width=\linewidth]{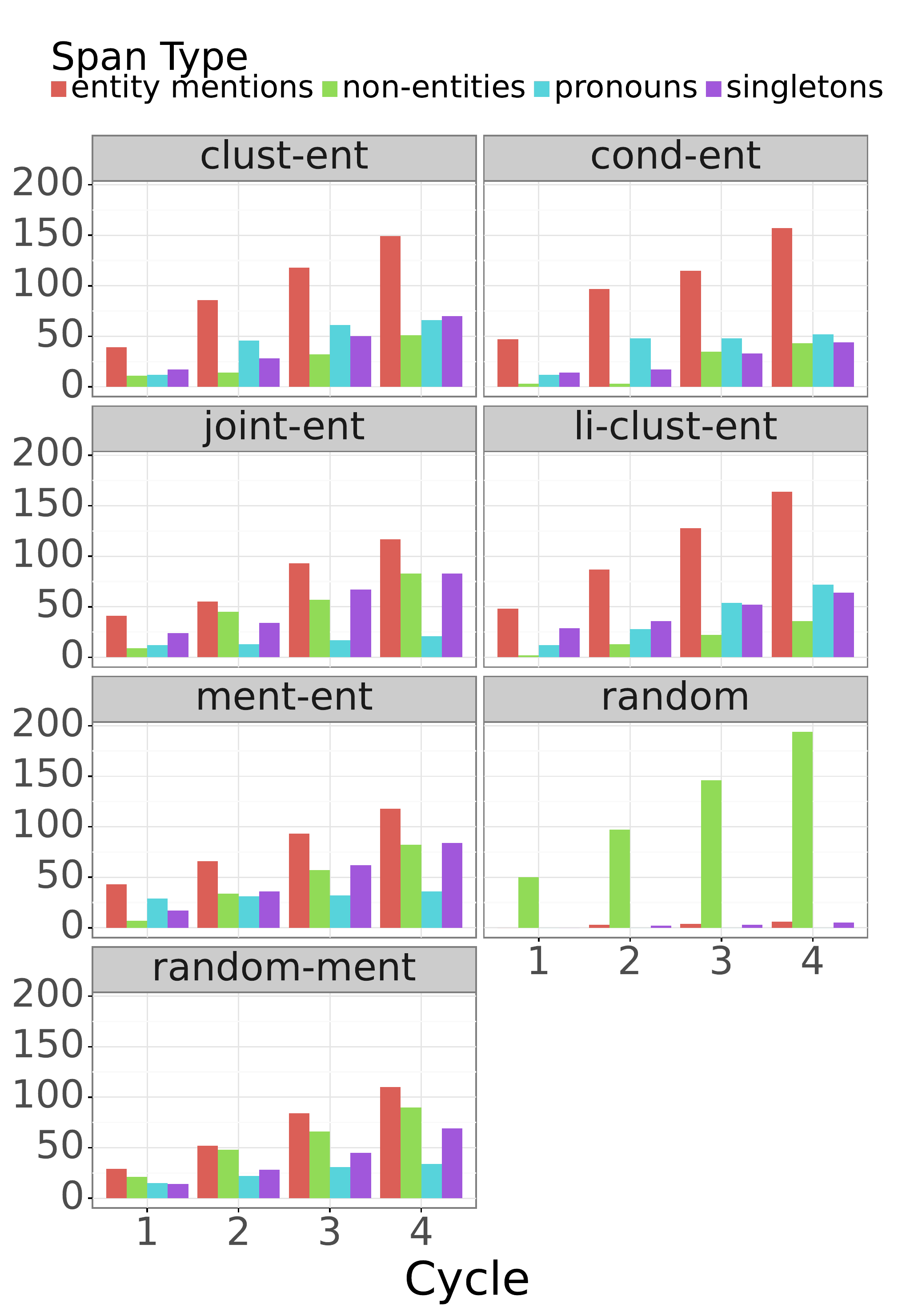}
    \caption{Cumulative counts of entities, non-entities, pronouns, and
    singletons
    sampled for each strategy over first four cycles of the \preco{} simulation. \textbf{Random} mostly samples
    non-entities. \textbf{Li-clust-ent} and \textbf{cond-ent} sample many
    entity mentions but avoid singletons.
    }
    \label{fig:preco_samples}
\end{figure}

\paragraph{Datasets}

\ontonotes~{\small 5.0} is the most common dataset for training and evaluating
\coref{}~\citep{pradhan-2013}.
The dataset contains news articles
and telephone conversations. Only non-singletons are annotated.
Our experiments transfer a model trained on \ontonotes{} to two
target datasets: \preco{} and \qbcoref{}.
\preco{} is a large corpus of grade-school reading comprehension
texts~\citep{chen-2018-preco}.
Unlike \ontonotes{}, \preco{} has annotated singletons. There are 37K training, 500 validation, and 500 test
documents.
Because the training set is so large, \citet{chen-2018-preco} only analyze
subsets of 2.5K documents.
Likewise, we reduce the training set to a subset of 2.5K
documents, comparable to the size of \ontonotes{}.

The \qbcoref{} dataset~\citep{guha-2015} contains trivia questions from Quizbowl
tournaments
that are densely packed with
entities from academic topics.
Like \preco{}, singletons are
annotated.
Unlike other datasets, the syntax is idiosyncratic and world knowledge is needed
to solve coreference.
Examples are pronouns before the first mention of named
entities and oblique references like ``this polity'' for
``the Hanseatic League''.
These complicated structures rarely occur in everyday text but serve
as challenging examples for \coref{}.
There are 240 training, 80
validation, and 80 test documents.

\subsection{Simulation: Uncertainty Sampling}
\label{ssec:sampling_results}

To compare different sampling
strategies, we first run experiments on \preco{}. We sample fifty spans from one document for each cycle. By the end
of a simulation run, 300 spans are sampled from six documents. For this
configuration, uncertainty sampling strategies generally reach higher accuracy
than the random baselines (Figure~\ref{fig:preco_simple}),
but \textbf{cond-ent} and \textbf{li-clust-ent} are worse than \textbf{random-ment}.

\subsubsection{Distribution of Sampled Span Types}
\label{ssec:distribution}
To understand the type of spans being sampled, we count entity
mentions,
non-entities, pronouns, and singletons that are sampled by each strategy
(Figure~\ref{fig:preco_samples}). \textbf{Random} samples very few
entities, while other strategies sample more entity mentions. \textbf{Clust-ent} and
\textbf{cond-ent} sample more entity mentions and pronouns because the sampling objective
prioritizes mentions that are difficult to link.
\textbf{Clust-ent}, \textbf{joint-ent}, and \textbf{ment-ent} sample more
singleton mentions. These strategies also show higher \avgfone{}
(Figure~\ref{fig:preco_simple}). For transferring from \ontonotes{} to \preco{},
annotating singletons is useful because only non-singleton mentions are labeled
in \ontonotes{}.  We notice \textbf{ment-ent} sampling pronouns, which
should obviously be entity mentions, only in the first cycle.
Many pronouns in \ontonotes{} are singletons, so the mention detector has trouble
distinguishing them initially in \preco{}.

\begin{figure}[!t]
    \centering
    \includegraphics[width=\linewidth]{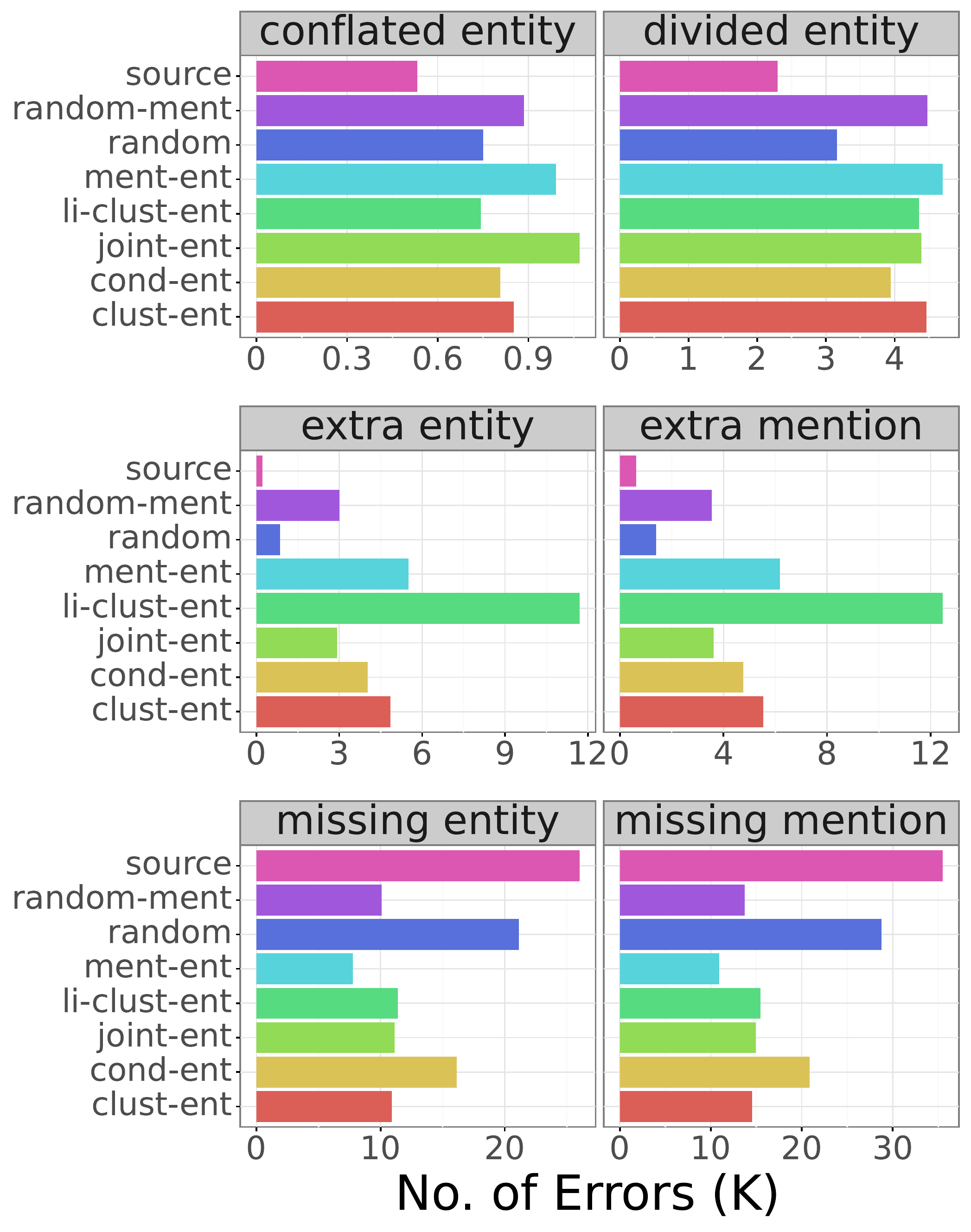}
    \caption{For each sampling strategy, we analyze the model from the last
    cycle of its \preco{} simulation. We compare the number of errors across common error
    types in \coref{}. The \textbf{source} \ontonotes{} model
    severely suffers from \emph{missing entities} and \emph{missing mentions}.
    \textbf{Ment-ent} helps most with reducing these errors.
    }
    \label{fig:preco_error}
\end{figure}

\subsubsection{Error Analysis}
\label{ssec:error}
\citet{kummerfeld-2013} enumerate the ways \coref{} models can go wrong:
\emph{missing entity}, \emph{extra entity}, \emph{missing mention}, \emph{extra
mention}, \emph{divided entity},
and \emph{conflated entity}. \emph{Missing entity} means a gold entity
cluster is missing. \emph{Missing mention} means a mention span for a gold
entity cluster is missing.  The same definitions apply for \emph{extra entity}
and \emph{extra mention}.
\emph{Divided entity} occurs when the model splits a gold entity cluster into
multiple ones. \emph{Conflated entity} happens when the model merges
gold entity clusters. For each strategy, we analyze the
errors of its final model from the
simulation's last cycle (Figure~\ref{fig:preco_error}).
We compare against the
\textbf{source} model that is only trained on \ontonotes{}.

The \textbf{source}
model makes many \emph{missing entity} and \emph{missing mention} errors.
It does not detect several entity spans in \preco{}, like
locations (``Long Island'') or ones spanning multiple words (``his kind
acts of providing everything that I needed'').
These spans are detected by uncertainty sampling strategies and \textbf{rand-ment}. \textbf{Ment-ent} is most effective at reducing ``missing''
errors. It detects gold entity clusters like
``constant communication'' and ``the best educated guess about the storm''.
By training on spans that confuse the mention detector, the model adapts to the
new domain by understanding what constitutes as an entity mention.

Surprisingly, \textbf{li-clust-ent} makes at least twice as many \emph{extra entity} and
\emph{extra mention}
errors than any other strategy. For the sentence, ``Living in a large building with only 10 bedrooms'', the
gold data identifies two entities: ``a large building with only 10
bedrooms'' and ``10 bedrooms''.
In both \ontonotes{} and \preco{}, the guidelines only allow the longest noun
phrase to be annotated.
Yet, the \textbf{li-clust-ent} model predicts additional mentions,
``a large building'' and ``only 10 bedrooms''.
We find that \textbf{li-clust-ent} tends to sample nested
spans (Table~\ref{tab:examples_preco}). Due to the summed entropy computation, nested spans
share similar values for clustered entropy as they share similar
antecedent-linking probabilities. This causes the \emph{extra entity} and
\emph{extra mention} errors because the model predicts there are additional
entity mentions within a mention span.

Finally, we see a stark difference between \textbf{random-ment} and \textbf{random}.
Out of all the sampling strategies, \textbf{random} is least effective at preventing \emph{missing entity} and \emph{missing mention} errors.
We are more likely to sample non-entities if we randomly sample from all spans in the document (Appendix~\ref{ssec:examples}).
By limiting the sampling pool to only spans that are likely to be entity mentions, we sample more spans that are useful to label for \coref{}.  Thus, the mention detector from neural models should be deployed during active learning.

\subsection{User Study: Reading and Labeling}
\label{ssec:human_labeling}

We hold a user study to observe the trade-off between reading and labeling.
Three annotators, with minimal \abr{nlp} knowledge, label spans
sampled from \preco{}. We use \textbf{ment-ent} to sample spans because the
strategy shows highest \avgfone{} (Figure~\ref{fig:preco_simple}).
First, the users read instructions (Appendix~\ref{ssec:user_appendix}) and practice labeling for
ten minutes.
Then, they complete two sessions: \textbf{FewDocs} and \textbf{ManyDocs}.
In each session, they label as much as possible for at least twenty-five
minutes.
In \textbf{FewDocs}, they read fewer documents and label roughly
seven spans per document.
In \textbf{ManyDocs}, they read more documents and label
about one span per document.

For labeling coreference, we develop a user interface that is open-sourced
(Figure~\ref{fig:ui}).
 To label the
 antecedent of
 the highlighted span, the user clicks on a contiguous span of tokens.
 The interface suggests
 overlapping candidates based on the spans that are retained by the \coref{}
 model.

\begin{figure}[!t]
\centering
\includegraphics[width=\linewidth]{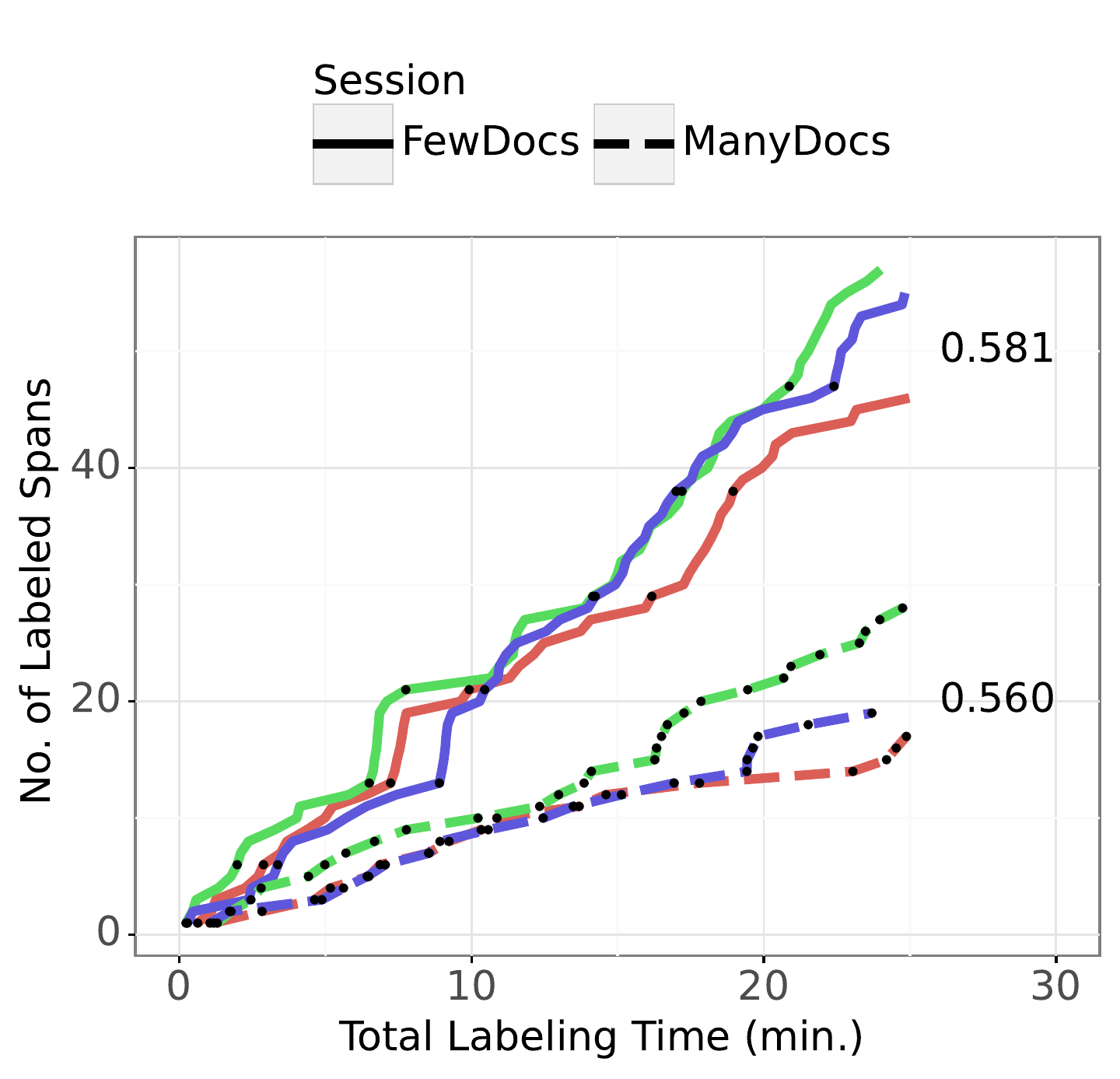}
\caption{The number of spans labeled within twenty-five minutes.
Each color indicates one of three users and the linetype
    designates the session.
Black dots mark the first span labeled in a different
    document. The mean \avgfone{} across users for each session is
    on the right. By restricting the number of read documents in
    \textbf{FewDocs}, users label at least twice as many spans and the model
    slightly
    improves in \avgfone{}. }
\label{fig:user}
\end{figure}

In the user study, participants label at least twice as much
in \textbf{FewDocs} compared to \textbf{ManyDocs} (Figure~\ref{fig:user}).
By labeling more spans in \textbf{FewDocs}, the mean \avgfone{} score is also
slightly higher.
Our findings show that the number
of read documents should be constrained to increase labeling
throughput.
Difference in number of labeled spans between \textbf{FewDocs}
and \textbf{ManyDocs} is more pronounced when two annotators volunteer to continue labeling after required duration
(Appendix~\ref{ssec:user_appendix}).

\begin{figure}[!t]
    \centering
    \begin{subfigure}[t]{\linewidth}
        \centering
        \includegraphics[width=0.90\linewidth]{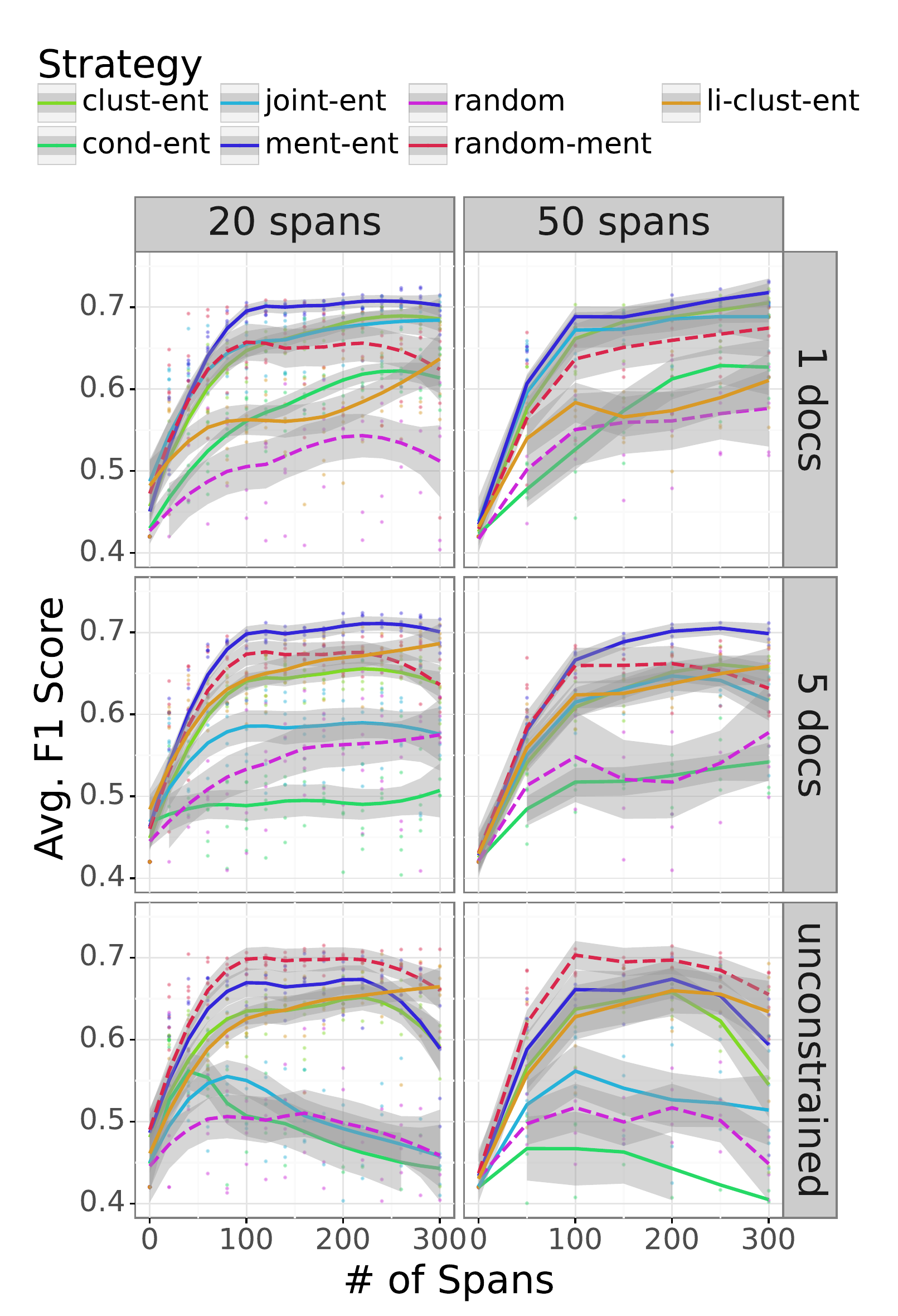}
        \caption{\preco{}}
        \label{fig:preco}
    \end{subfigure}
    \begin{subfigure}[t]{\linewidth}
        \centering
        \includegraphics[width=0.80\linewidth]{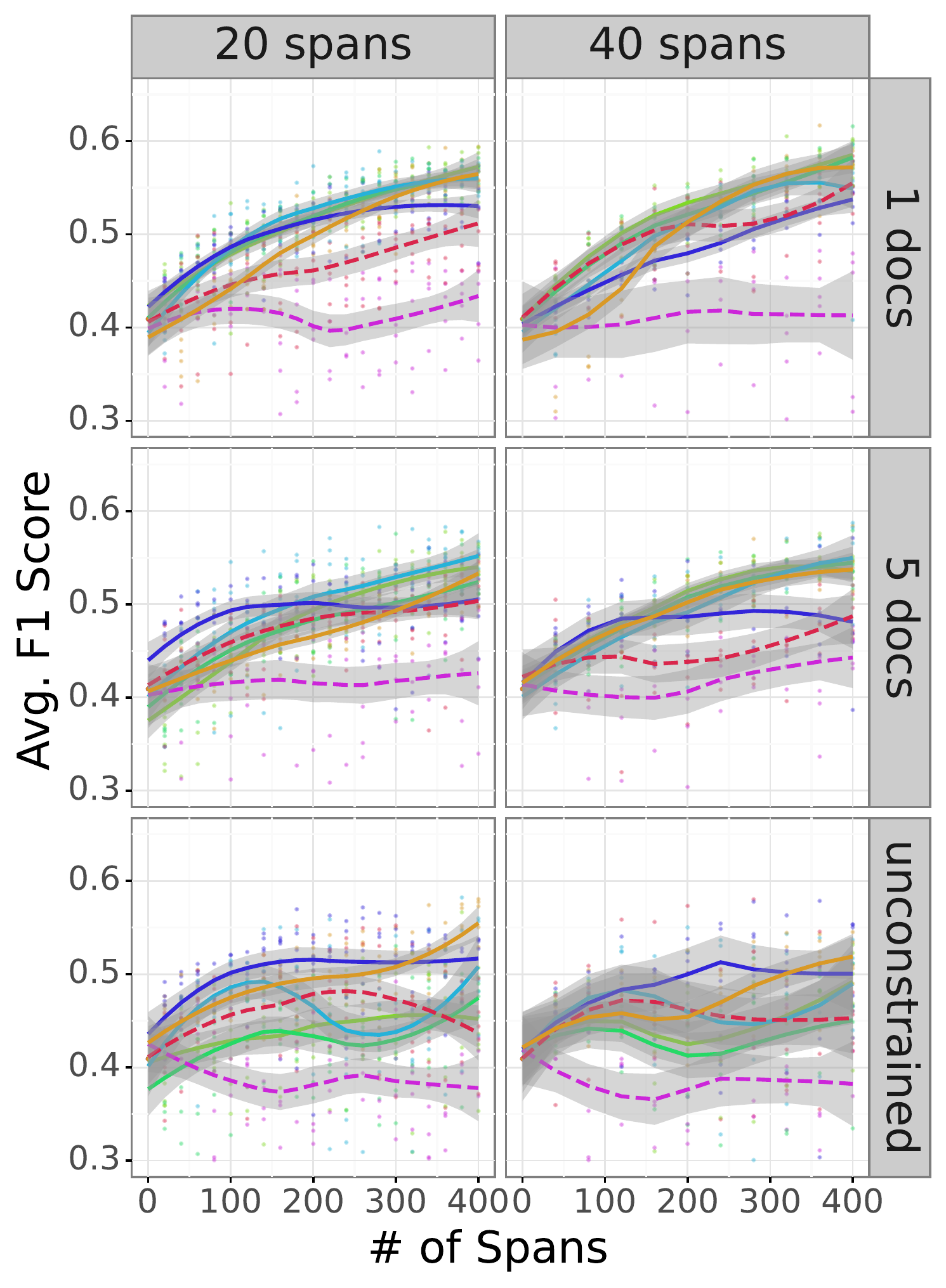}
        \caption{\qbcoref{}}
        \label{fig:qbcoref}
    \end{subfigure}
    \caption{Test \avgfone{} on \preco{} and \qbcoref{} of each strategy
    throughout simulations. Each row varies in~$m$, the maximum number of
    documents read per cycle. Each column varies in $k$, the number of annotated
    spans per cycle.
    For~$m$ of one or five, \textbf{ment-ent} shows highest \avgfone{}
    for \preco{} and other uncertainty sampling strategies are best for
    \qbcoref{}. When $m$ is unconstrained, many strategies show unstable
    training.
    }
    \label{fig:f1}
\end{figure}

\subsection{Simulation: Uncertainty Sampling and Reading-Labeling Trade-off}
\label{ssec:adaptation}

We finally run simulations to explore \textit{both} sources of model uncertainty and the
trade-off between reading and labeling. The earlier experiments have individually
looked at each aspect.
Now, we analyze the interaction between both factors to understand which
combination works best for adapting \coref{} to new domains.
We run simulations on \preco{} and \qbcoref{} that trade-off the number
of documents read~$m$ with the number of annotated spans~$k$
(Figure~\ref{fig:f1}).  We vary~$m$ between one, five, and an unconstrained number of documents.
For \preco{}, we set~$k$ to twenty and fifty.
For \qbcoref{}, we set~$k$ to twenty and forty.
These results are also presented in numerical form (Appendix~\ref{ssec:num}).

\paragraph{\preco{}}
For \preco{}, the test \avgfone{} of
\icoref{} trained on the full training dataset is 0.860.
When $m$ is constrained to one or five, \avgfone{} can reach around 0.707 from
training the model on only 300 spans sampled by
\textbf{ment-ent}.
As $m$ increases, fewer spans are sampled per document and
all sampling strategies deteriorate.
After training on sparsely annotated documents, the model tends to predict
singletons rather than cluster coreferent spans.
Like in the user study, we see
benefits when labeling more spans within a document.
Interestingly, \textbf{li-clust-ent} performs better when document
reading is not constrained to one document.
The issue with
\textbf{li-clust-ent} is that it samples nested mention spans
(Section~\ref{ssec:error}).
Duplicate sampling is less severe if spans can be sampled across more
documents. Another strategy that suffers from duplicate sampling is
\textbf{cond-ent} because it mainly samples pronouns. For some documents, the
pronouns all link to the same entity cluster.
As a result, the model trains on a
less diverse set of entity mentions and \textbf{cond-ent} drops in
\avgfone{} as the simulation continues.

\paragraph{\qbcoref{}}
For \qbcoref{}, the
test \avgfone{} of \icoref{} trained on the full training dataset is 0.795.
When we constrain $m$ to one or five, \textbf{li-clust-ent}, \textbf{clust-ent}, \textbf{cond-ent}, and
\textbf{joint-ent} have high \avgfone{}. Clustering entity mentions in
\qbcoref{} questions is difficult, so these strategies
help target ambiguous mentions (Table~\ref{tab:examples_qbcoref}). \textbf{Ment-ent} is less useful
because demonstratives are abundant in \qbcoref{} and
make mention detection easier.
\textbf{Li-clust-ent} still
samples nested entity mentions, but annotations for these spans help clarify
interwoven entities in Quizbowl questions. Unlike \preco{},
\textbf{li-clust-ent} does not sample duplicate entities because nested entity
mentions belong to different clusters and need to be distinguished.

Overall, the most helpful strategy depends on the domain.
For domains like
\preco{} that contain long documents with many singletons, \textbf{ment-ent} is
useful.
For domains like \qbcoref{} where resolving coreference is difficult,
we need to target linking uncertainty.
Regardless of the dataset,
\textbf{random} performs worst.
\textbf{Random-ment} has much higher \avgfone{}, which shows the importance
of the mention detector in active learning.
Future work should
determine the appropriate strategy for a given domain and annotation
setup.

\section{Related Work}

\citet{gasperin-2009} present the first work on active learning for \coref{}
yet observe negative results:
active learning is not more effective
than random sampling.
\citet{miller-2012} explore
different settings for labeling \coref{}.
First, they
label the most uncertain pairs of spans in the corpus.
Second, they label all pairs in the most
uncertain documents.
The first approach beats random sampling but requires
the annotator to infeasibly read many documents.
The second approach is more realistic but loses to random sampling.
\citet{zhao-2014} argue that active learning helps domain adaptation of
\coref{}.
\citet{sachan-2015} treat
pairwise
annotations as optimization constraints.
\citet{li-2020} replace pairwise annotations with
discrete annotations and experiment active learning with neural models.

Active learning has been exhaustively studied for text
classification~\citep{lewis-1994,zhu-2008,zhang-2017}.
Text classification is a much simpler task, so researchers investigate  strategies beyond uncertainty sampling.
\citet{yuan-2020-alps} use language model surprisal to cluster documents and then
sample representative points for each cluster.
\citet{margatina-2021} search for constrastive examples, which are documents
that are similar in
the feature space yet differ in predictive likelihood.
Active learning is also applied to tasks like
machine
translation~\citep{liu-2018}, visual question
answering~\citep{karamcheti-2021}, and entity alignment~\citep{bing-2021}.

Rather than solely running simulations, other papers have also ran
user studies or developed user-friendly interfaces. \citet{wei-2019} hold a user study for active
learning to observe the time to annotate clinical named entities.
\citet{lee-2020} develop active learning for
language learning that adjusts labeling difficulty based on user
skills. \citet{klie-2020} create a human-in-the-loop pipeline to improve
entity linking for low-resource domains.

\section{Conclusion}

Neural \coref{} models desparately depend on large, labeled data.
 We use active learning to transfer a model trained on \ontonotes{}, the
 ``de facto'' dataset, to new domains.
Active learning for \coref{} is difficult because the problem does not only
concern sampling examples. We must consider different aspects, like sources of
model uncertainty and cost of reading documents.
Our work explores these factors through exhaustive simulations.
Additionally, we develop a user interface to run a user study from which we observe human annotation time and
throughput.
In both simulations and the user study, \coref{} improves from
continued training on spans sampled from the same document rather than different
contexts.
Surprisingly, sampling by entropy in mention detection, rather than linking, is most helpful for domains like
\preco{}. This opposes the assumption that the uncertainty strategy must
be directly tied to the training objective.
Future work may extend our contributions to multilingual transfer or
multi-component tasks, like open-domain \abr{qa}.

\section{Ethical Considerations}

This paper involves a user study to observe the trade-off between reading and labeling costs for annotating coreference.
The study has been approved by \abr{irb} to collect data about human behavior.
Any personal information will be anonymized prior to paper submission or publication.
All participants are fully aware of the labeling task and the information that
will be collected from them. They are appropriately compensated for their
labeling efforts.

\section*{Acknowledgements}

We thank Ani Nenkova, Jonathan Kummerfeld, Matthew Shu, Chen Zhao, and the anonymous reviewers for their insightful feedback.
We thank the user study participants for supporting this work through annotating
data.
Michelle Yuan and Jordan Boyd-Graber are supported in part by Adobe Inc.
Any opinions, findings, conclusions, or recommendations
expressed here are those of the authors and do not necessarily reflect the view of the sponsors.

\bibliography{bib/journal-full,bib/jbg,bib/michelle}
\bibliographystyle{style/acl_natbib}
\clearpage
\appendix
\section{Appendix}
\label{sec:appendix}

\subsection{Coreference Resolution Models}
\label{ssec:neural}

\paragraph{\ctof}

In \ctof{},
a pairwise scorer computes $s(x,y)$ to learn antecedent distribution $P(Y)$
(Equation~\ref{eq:antecedent}).
The model's pairwise scorer judges whether span $x$ and span $y$ are
coreferent based on their antecedent score $s_a$ and individual mention scores
$s_m$,
\begin{equation}
    s(x,y) =
        \begin{cases}
            0  & y=\epsilon\\
            s_m(x) + s_m(y) + s_a(x,y) & y \neq \epsilon
        \end{cases},
    \label{eq:pairwise}
\end{equation}
Suppose $\bm{g_x}$ and $\bm{g_y}$ are the span representations of $x$ and $y$,
respectively.
Mention scores and antecedent scores are then computed with feedforward networks
$FFNN_m$ and $FFNN_c$,
\begin{align}
    \label{eq:unary}
s_m(x) &= FFNN_m(\bm{g_x}) \\
s_a(x,y) &= FFNN_a(\bm{g_x}, \bm{g_y}, \phi(x,y)).
\end{align}
The input $\phi(x,y)$ includes features like the distance between spans.
The unary mention score $s_m$ can be viewed as the likelihood that the span is
an entity mention. For computational purposes, the \ctof{} model only retains
top-$k$ spans with the highest unary mention scores. \citet{lee-2018} provide more details about the pairwise scorer and span pruning.

\paragraph{Incremental Clustering}
We elaborate upon the clustering algorithm of \icoref{} here.
As the algorithm processes spans in the
document, each span is either placed in a cluster from $\mathcal{C}$ or
added to a new cluster.
To learn the distribution over clusters (Equation~\ref{eq:cluster}), the algorithm first creates a cluster
representation $\bm{g_c}$ that is an aggregate of span
representation that is an aggregate of span
representations over spans that currently exist in the cluster.
(Equation~\ref{eq:cluster_rep}).
With cluster and span representations, individual spans and entity clusters are
mapped into a shared space. Then, we can compute $s(x,c)$ using the same
pairwise scorer as \citet{lee-2018}.
Suppose that model predicts $c^*$ as most likely cluster: $c^* = \argmax_{c \in \mathcal{C}}
s(x,c)$. Now, the algorithm makes one of two decisions:
\begin{enumerate}
\item If $s(x, c^*) > 0$, then $x$ is assigned to $c^*$ and
    update $\bm{g_{c^*}}$ such that
    \begin{equation}
        \bm{g_{c^*}} = s_e(c^*,x) \bm{g_{c^*}} + (1-s_e(c^*,x)) \bm{g_x},
        \label{eq:cluster_rep}
    \end{equation}
    where $s_e$ is a learned weight.
\item If $s(x,c^*) \le 0$, then a new entity cluster $c_x = \{x\}$ is added to
$\mathcal{C}$.
\end{enumerate}
The algorithm repeats for each span in the document.

Like \ctof{}, the \icoref{} model only retains top-$k$ spans with highest unary
mention score.  All of our active learning baselines
(Section~\ref{sec:experiments}), except \textbf{random}, sample spans
from this top-$k$ pool of spans.

\subsection{Training Configuration}
\label{ssec:params}
The \spanbertlarge{} encoder has 334M parameters and \icoref{} has 373M parameters in total. For model fine-tuning, we train for a maximum of fifty epochs and implement early stopping
with a patience of ten epochs. We set
top span pruning to 0.4, dropout to 0.4, gradient clipping to 10.0, and learning
rate to 1e-4 for Adam optimizer.
The hyperparameter configuration is based on
results from prior work~\citep{lee-2017-coref,xia-2020}.

All experiments in the paper are ran on NVIDIA Tesla V100 GPU and 2.2 GHz Intel
Xeon Silver 4114 CPU processor.

\subsection{Simulation Time}
\label{ssec:sim_time}

\begin{table}[t]
    \centering
\begin{tabular}{lrr}
    \toprule
    Strategy & \preco & \qbcoref \\
    \midrule
    \textbf{random} & 2 &$<1$ \\
\textbf{random-ment} & 4 &$<1$\\
\textbf{ment-ent} & 5 &$<1$\\
\textbf{li-clust-ent} & 12 & $<1$ \\
\textbf{clust-ent} & 12 & 1 \\
\textbf{cond-ent} & 14 & 1 \\
\textbf{joint-ent} & 16 & 1 \\
\bottomrule
\end{tabular}
    \caption{The time (minutes) to sample a batch of fifty spans from five documents from
    either \preco{} or \qbcoref{} for a
    given active learning strategy. On large datasets like \preco{}, we see that
    \textbf{li-clust-ent}, \textbf{clust-ent}, \textbf{cond-ent}, and \textbf{joint-ent} are slower
    because the strategy needs to incrementally cluster each span and then
    compute clustering entropy.}
\label{tab:sample_time}
\end{table}

We compare the time to sample fifty spans between different active learning strategies
for \preco{} and \qbcoref{} (Table~\ref{tab:sample_time}).  For \preco{},
\textbf{clust-ent}, \textbf{cond-ent}, and \textbf{joint-ent} are slower
because they need to run documents through \icoref{} and get span-cluster
likelihood. On the other hand, \textbf{ment-ent} only needs unary scores $s_m$,
which is much faster to compute.  Thus, for both datasets, running
\textbf{ment-ent} takes about the same time as \textbf{random-ment}.

For \qbcoref{}, fine-tuning \icoref{} on fifty spans takes three minutes and
fine-tuning on full
training set takes thirty-four minutes.
For \preco{}, fine-tuning \icoref{} on fifty spans takes nine minutes and
fine-tuning on full
training set takes five hours and 22 minutes.

\begin{figure}[!t]
    \centering
    \begin{subfigure}[!t]{\linewidth}
        \centering
        \includegraphics[width=0.95\linewidth]{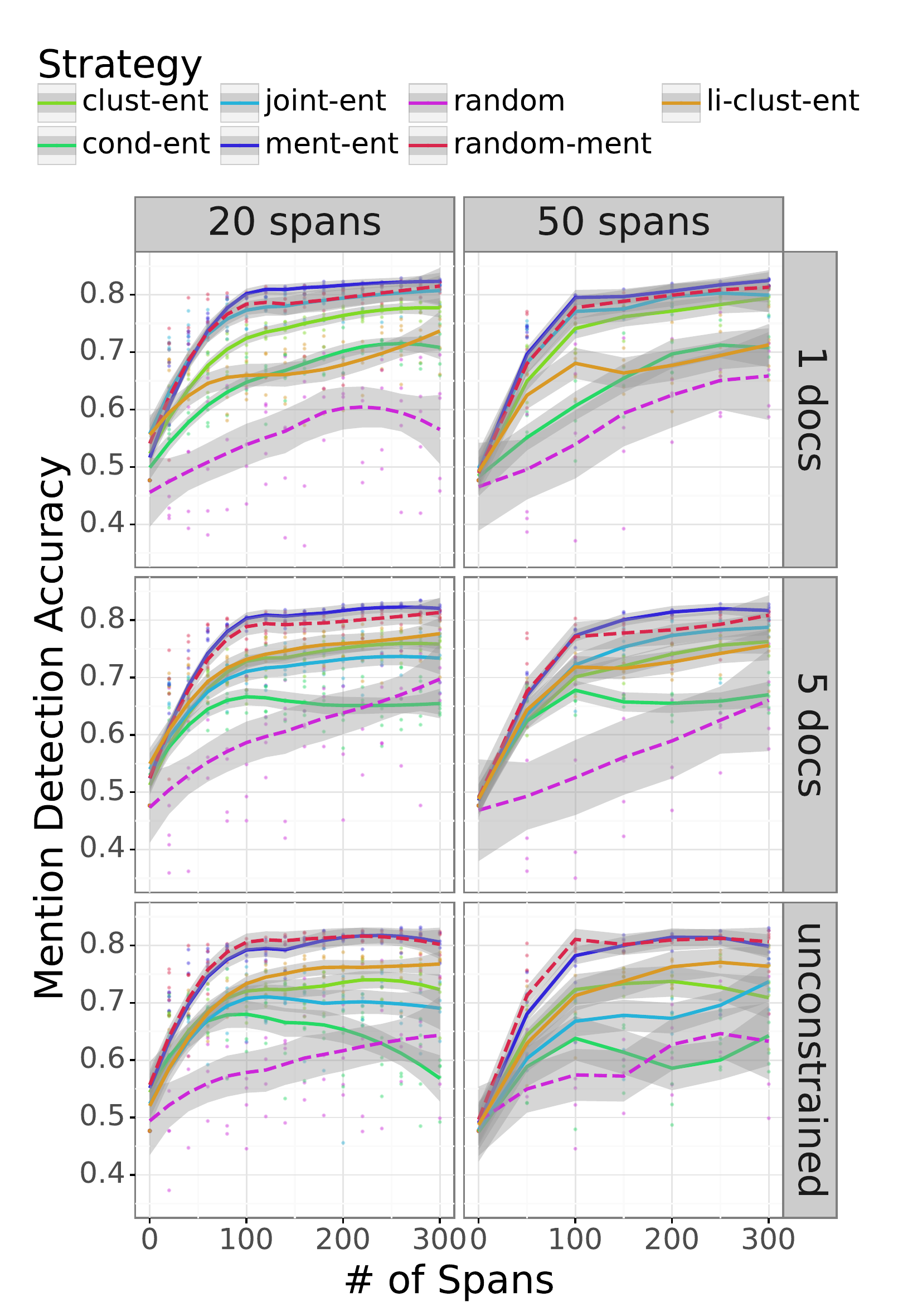}
        \caption{\preco{}}
        \label{fig:preco_ment}
    \end{subfigure}
    \begin{subfigure}[!t]{\linewidth}
        \centering
        \includegraphics[width=0.88\linewidth]{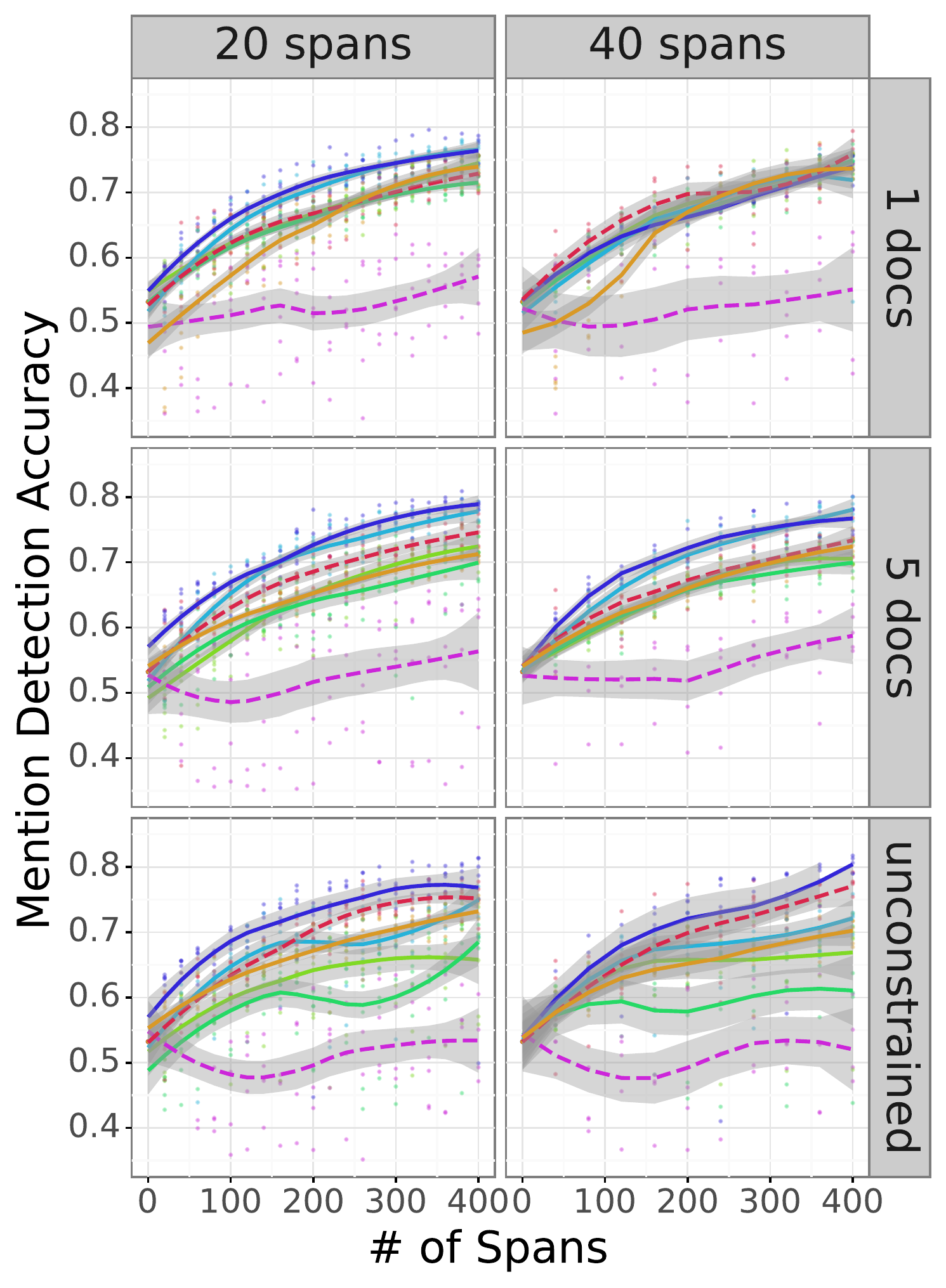}
        \caption{\qbcoref{}}
        \label{fig:qbcoref_ment}
    \end{subfigure}
    \caption{Comparing mention detection accuracy on test set for different active learning strategies across reading/labeling
    configurations. The plots are formatted in the same way as Figure~\ref{fig:f1}.
    Generally, mention detection improves most from \textbf{ment-ent} sampling.}
    \label{fig:detection}
\end{figure}

\subsection{Mention Detection Accuracy}
\label{ssec:detection}

For the annotation simulation in Section~\ref{sec:experiments}, we also record
mention detection accuracy. As \textbf{ment-ent} targets ambiguity in mention
detection, it is the most effective strategy for improving mention detection
(Figure~\ref{fig:detection}). The strategy is unaffected by labeling setup
parameters, like the number of spans labeled per cycle or the number of
documents read per cycle. For strategies like \textbf{cond-ent} and
\textbf{joint-ent}, mention detection accuracy is stagnant or decreases as more
spans are sampled (Figure~\ref{fig:preco_ment}).  Due to deteriorating mention detection, the \avgfone{} of
models also drop.

\subsection{Numerical Results}
\label{ssec:num}
The results for \avgfone{} and mention detection accuracy are presented as
graphs throughout the paper. To concretely understand the differences between
the methods, we provide
results in numerical form (Tables~\ref{tab:preco_num},\ref{tab:qb_num}). We show
results from the \preco{} and \qbcoref{} simulations where twenty spans are labeled each
cycle and the number of documents read is either one or an unconstrained amount. The values in the tables show the mean and variance of \avgfone{} and mention detection accuracy over five different runs.

\begin{table*}
    \centering
\begin{tabular}{lllll}
\toprule
    Total No.~of Labeled Spans & $m$ & Strategy & \avgfone{} & Mention Accuracy \\
\midrule
100 & 1 & clust-ent &  0.64 $\pm$ 0.02 &  0.71 $\pm$ 0.03 \\
    &               & cond-ent &  0.57 $\pm$ 0.02 &  0.66 $\pm$ 0.02 \\
    &               & joint-ent &  0.64 $\pm$ 0.03 &  0.76 $\pm$ 0.02 \\
    &               & ment-ent &  \textbf{0.70 $\pm$ 0.01} &  \textbf{0.80 $\pm$ 0.00} \\
    &               & random &  0.43 $\pm$ 0.09 &  0.49 $\pm$ 0.11 \\
    &               & random-ment &  0.65 $\pm$ 0.04 &  0.78 $\pm$ 0.02 \\
    &               & li-clust-ent &  0.56 $\pm$ 0.02 &  0.65 $\pm$ 0.03 \\
    \cline{2-5}
    & unconstrained & clust-ent &  0.62 $\pm$ 0.03 &  0.70 $\pm$ 0.03 \\
    &               & cond-ent &  0.43 $\pm$ 0.09 &  0.67 $\pm$ 0.04 \\
    &               & joint-ent &  0.55 $\pm$ 0.06 &  0.71 $\pm$ 0.05 \\
    &               & ment-ent &  0.65 $\pm$ 0.03 &  0.76 $\pm$ 0.03 \\
    &               & random &  0.48 $\pm$ 0.07 &  0.54 $\pm$ 0.07 \\
    &               & random-ment &  \textbf{0.69 $\pm$ 0.01} &  \textbf{0.80 $\pm$ 0.01} \\
    &               & li-clust-ent &  0.62 $\pm$ 0.01 &  0.73 $\pm$ 0.01 \\
    \cline{1-5}
200 & 1 & clust-ent &  0.68 $\pm$ 0.01 &  0.77 $\pm$ 0.01 \\
    &               & cond-ent &  0.62 $\pm$ 0.02 &  0.70 $\pm$ 0.03 \\
    &               & joint-ent &  0.68 $\pm$ 0.03 &  0.80 $\pm$ 0.02 \\
    &               & ment-ent &  \textbf{0.71 $\pm$ 0.01} &  \textbf{0.82 $\pm$ 0.00} \\
    &               & random &  0.48 $\pm$ 0.18 &  0.55 $\pm$ 0.21 \\
    &               & random-ment &  0.65 $\pm$ 0.05 &  0.77 $\pm$ 0.07 \\
    &               & li-clust-ent &  0.57 $\pm$ 0.05 &  0.67 $\pm$ 0.04 \\
    \cline{2-5}
    & unconstrained & clust-ent &  0.65 $\pm$ 0.02 &  0.73 $\pm$ 0.03 \\
    &               & cond-ent &  0.36 $\pm$ 0.08 &  0.63 $\pm$ 0.07 \\
    &               & joint-ent &  0.40 $\pm$ 0.12 &  0.67 $\pm$ 0.12 \\
    &               & ment-ent &  0.67 $\pm$ 0.03 &  \textbf{0.81 $\pm$ 0.01} \\
    &               & random &  0.49 $\pm$ 0.08 &  0.61 $\pm$ 0.07 \\
    &               & random-ment &  \textbf{0.69 $\pm$ 0.01} &  \textbf{0.81 $\pm$ 0.00} \\
    &               & li-clust-ent &  0.65 $\pm$ 0.03 &  0.75 $\pm$ 0.03 \\
    \cline{1-5}
300 & 1 & clust-ent &  0.68 $\pm$ 0.02 &  0.78 $\pm$ 0.01 \\
    &               & cond-ent &  0.61 $\pm$ 0.03 &  0.70 $\pm$ 0.04 \\
    &               & joint-ent &  \textbf{0.69 $\pm$ 0.02} &  0.81 $\pm$ 0.01 \\
    &               & ment-ent &  \textbf{0.69 $\pm$ 0.02} &  \textbf{0.82 $\pm$ 0.00} \\
    &               & random &  0.50 $\pm$ 0.09 &  0.58 $\pm$ 0.10 \\
    &               & random-ment &  0.61 $\pm$ 0.10 &  0.81 $\pm$ 0.01 \\
    &               & li-clust-ent &  0.63 $\pm$ 0.05 &  0.73 $\pm$ 0.05 \\
    \cline{2-5}
    & unconstrained & clust-ent &  0.51 $\pm$ 0.12 &  0.70 $\pm$ 0.04 \\
    &               & cond-ent &  0.33 $\pm$ 0.07 &  0.57 $\pm$ 0.04 \\
    &               & joint-ent &  0.41 $\pm$ 0.05 &  0.69 $\pm$ 0.04 \\
    &               & ment-ent &  0.54 $\pm$ 0.07 &  0.80 $\pm$ 0.02 \\
    &               & random &  0.40 $\pm$ 0.04 &  0.60 $\pm$ 0.13 \\
    &               & random-ment &  0.65 $\pm$ 0.05 &  \textbf{0.80 $\pm$ 0.04} \\
    &               & li-clust-ent &  \textbf{0.67 $\pm$ 0.02} &  0.78 $\pm$ 0.01 \\
\bottomrule
\end{tabular}
    \caption{Results of \preco{} simulation in numerical form, accompanying the
    graphs in Figures~\ref{fig:preco} and~\ref{fig:preco_ment}. The table
    shows \avgfone{} and mention detection accuracy of experiments where twenty
    spans are sampled and labeled each cycle. Results are shown for $m$, the
    maximum number of documents read, equal to one and also unconstrained.
    }
\label{tab:preco_num}
\end{table*}

\begin{table*}
    \centering
\begin{tabular}{lllll}
\toprule
    Total No.~of Labeled Spans & $m$ & Strategy & \avgfone{} & Mention Accuracy \\
\midrule
100 & 1 & clust-ent &  0.47 $\pm$ 0.06 &  0.62 $\pm$ 0.06 \\
    &               & cond-ent &  0.47 $\pm$ 0.03 &  0.61 $\pm$ 0.03 \\
    &               & joint-ent &  \textbf{0.50 $\pm$ 0.03} &  \textbf{0.65 $\pm$ 0.02} \\
    &               & ment-ent &  \textbf{0.50 $\pm$ 0.01} &  0.66 $\pm$ 0.03 \\
    &               & random &  0.40 $\pm$ 0.07 &  0.53 $\pm$ 0.07 \\
    &               & random-ment &  0.44 $\pm$ 0.06 &  0.63 $\pm$ 0.04 \\
    &               & li-clust-ent &  0.45 $\pm$ 0.02 &  0.59 $\pm$ 0.03 \\
    \cline{2-5}
    & unconstrained & clust-ent &  0.41 $\pm$ 0.05 &  0.59 $\pm$ 0.07 \\
    &               & cond-ent &  0.39 $\pm$ 0.10 &  0.57 $\pm$ 0.05 \\
    &               & joint-ent &  0.50 $\pm$ 0.01 &  0.66 $\pm$ 0.02 \\
    &               & ment-ent &  \textbf{0.51 $\pm$ 0.02} &  \textbf{0.69 $\pm$ 0.01} \\
    &               & random &  0.36 $\pm$ 0.08 &  0.48 $\pm$ 0.10 \\
    &               & random-ment &  0.48 $\pm$ 0.02 &  0.65 $\pm$ 0.01 \\
    &               & li-clust-ent &  0.47 $\pm$ 0.01 &  0.62 $\pm$ 0.02 \\
    \cline{1-5}
200 & 1 & clust-ent &  0.52 $\pm$ 0.01 &  0.67 $\pm$ 0.01 \\
    &               & cond-ent &  0.52 $\pm$ 0.02 &  0.66 $\pm$ 0.02 \\
    &               & joint-ent &  \textbf{0.53 $\pm$ 0.03} &  0.70 $\pm$ 0.03 \\
    &               & ment-ent &  0.51 $\pm$ 0.02 &  \textbf{0.71 $\pm$ 0.02} \\
    &               & random &  0.40 $\pm$ 0.06 &  0.53 $\pm$ 0.08 \\
    &               & random-ment &  0.48 $\pm$ 0.05 &  0.68 $\pm$ 0.01 \\
    &               & li-clust-ent &  0.49 $\pm$ 0.01 &  0.64 $\pm$ 0.02 \\
    \cline{2-5}
    & unconstrained & clust-ent &  0.45 $\pm$ 0.04 &  0.64 $\pm$ 0.06 \\
    &               & cond-ent &  0.39 $\pm$ 0.06 &  0.55 $\pm$ 0.06 \\
    &               & joint-ent &  0.48 $\pm$ 0.05 &  \textbf{0.70 $\pm$ 0.03} \\
    &               & ment-ent &  0.49 $\pm$ 0.08 &  0.68 $\pm$ 0.13 \\
    &               & random &  0.34 $\pm$ 0.08 &  0.50 $\pm$ 0.11 \\
    &               & random-ment &  0.49 $\pm$ 0.04 &  \textbf{0.70 $\pm$ 0.01} \\
    &               & li-clust-ent &  \textbf{0.50 $\pm$ 0.03} &  0.68 $\pm$ 0.02 \\
    \cline{1-5}
300 & 1 & clust-ent &  0.54 $\pm$ 0.02 &  0.70 $\pm$ 0.02 \\
    &               & cond-ent &  \textbf{0.55 $\pm$ 0.02} &  0.70 $\pm$ 0.02 \\
    &               & joint-ent &  \textbf{0.55 $\pm$ 0.02} &  0.74 $\pm$ 0.01 \\
    &               & ment-ent &  0.53 $\pm$ 0.02 &  \textbf{0.75 $\pm$ 0.02} \\
    &               & random &  0.42 $\pm$ 0.05 &  0.55 $\pm$ 0.06 \\
    &               & random-ment &  0.49 $\pm$ 0.03 &  0.69 $\pm$ 0.03 \\
    &               & li-clust-ent &  0.53 $\pm$ 0.04 &  0.71 $\pm$ 0.02 \\
    \cline{2-5}
    & unconstrained & clust-ent &  0.46 $\pm$ 0.04 &  0.67 $\pm$ 0.06 \\
    &               & cond-ent &  0.42 $\pm$ 0.07 &  0.58 $\pm$ 0.12 \\
    &               & joint-ent &  0.43 $\pm$ 0.11 &  0.68 $\pm$ 0.08 \\
    &               & ment-ent &  0.50 $\pm$ 0.06 &  0.74 $\pm$ 0.04 \\
    &               & random &  0.34 $\pm$ 0.18 &  0.45 $\pm$ 0.23 \\
    &               & random-ment &  0.47 $\pm$ 0.08 &  \textbf{0.75 $\pm$ 0.02} \\
    &               & li-clust-ent &  \textbf{0.52 $\pm$ 0.03} &  0.71 $\pm$ 0.01 \\
\bottomrule
\end{tabular}
\caption{Results of \qbcoref{} simulation in numerical form, accompanying the
    graphs in Figures~\ref{fig:qbcoref} and~\ref{fig:qbcoref_ment}. The table
    shows \avgfone{} and mention detection accuracy of experiments where twenty
    spans are sampled and labeled each cycle. Results are shown for $m$, the
    maximum number of documents read, equal to one and also unconstrained.}
\label{tab:qb_num}
\end{table*}

\begin{figure*}[!t]
    \centering
    \includegraphics[width=\linewidth]{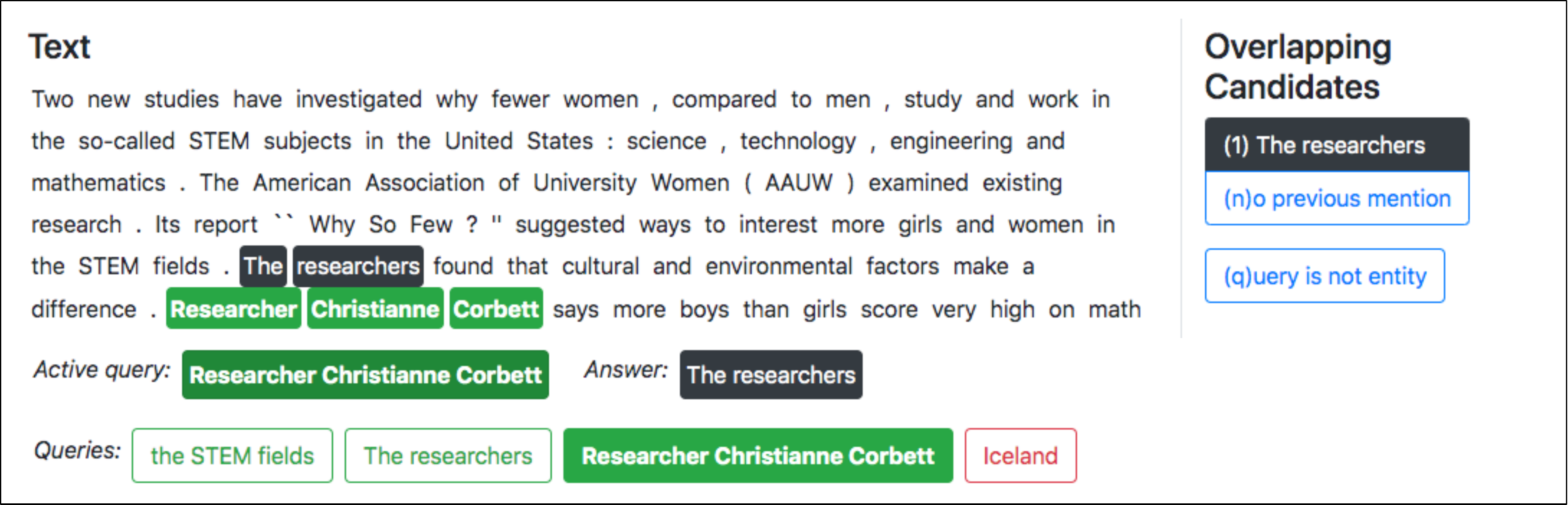}
    \caption{On the user interface, the sampled span
    is highlighted and the user must select an antecedent. If
    no antecedents exist or the span is not an entity mention, then the user
    will click the corresponding buttons.
    }
    \label{fig:ui}
\end{figure*}

\begin{figure}[!t]
    \centering
    \includegraphics[width=\linewidth]{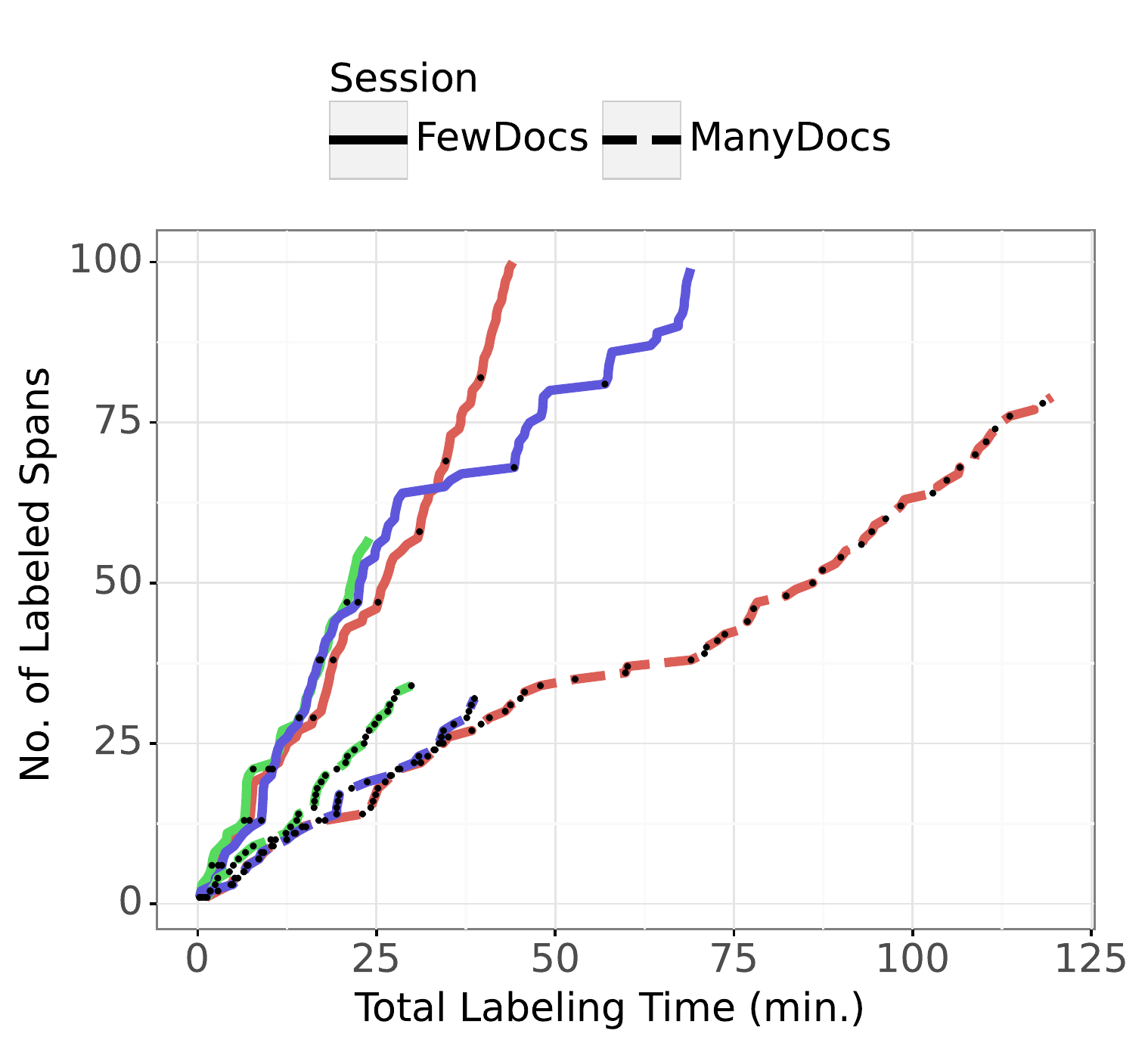}
    \caption{Full annotation times of participants (distinguished by color) during the user study. Over a
    longer period of time, the difference in number of labeled spans between the
    two sessions is much more pronounced. Within fourty-five minutes, the
    {\color{red}red} user can label a
    hundred spans in the \textbf{FewDocs} session but only labels about thirty spans
    in the \textbf{ManyDocs} session.}
    \label{fig:full}
\end{figure}

\subsection{User Study}
\label{ssec:user_appendix}

\paragraph{Instructions to Participants} We give the following instructions
to user study participants:
\begin{displayquote}
\normalsize
You will be shown several sentences from a
document. We have highlighted a mention (a word or phrase) of an entity (a
person, place, or thing). This entity mention may be a pronoun (such as ``she''
or ``their'') or something else. 

We need your help to find an earlier mention of
the same entity, whether in the same sentence or in an earlier sentence. The
mention does not have to be the immediately previous one. 

If the span is not an
entity mention or does not have an antecedent, please make note of it on the
interface.
\end{displayquote}

\paragraph{User Interface} We design a user interface for annotators to label
coreference (Figure~\ref{fig:ui}). The user interface takes the sampled spans from active learning as
input. Afterward, it will present the document and highlight the sampled spans
in the document. The user the proceeds to go through the list of ``Queries''.
For the ``Active query'', they need to either: find its antecedent, mark there
is ``no previous mention'', or indicate that ``query is not an entity''. The
interface will suggest some overlapping candidates to help narrow down the
user's search. The candidates are spans that the \coref{} model scores as likely entity mentions. Users may use keyboard shortcuts to minimize labeling time. The code for the user interface is released along with the code for the simulations.

\paragraph{Extending Annotation Time}

User study participants are asked to annotate at least twenty-five minutes
(Section~\ref{ssec:human_labeling}). During the study, two participants continue to label
after the minimum duration. Figure~\ref{fig:full} shows full results from the
user study. Over a longer duration, the differences between the \textbf{FewDocs}
and \textbf{ManyDocs} sessions are clearer.

\subsection{Examples of Sampled Spans}
\label{ssec:examples}
We provide examples of spans that are sampled from the experiments. For these
examples, we look at the simulation where document reading is constrained to one
document and twenty spans are sampled per cycle. We compare the spans sampled by
each strategy for both \preco{} (Table~\ref{tab:examples_preco}) and \qbcoref{}
(Table~\ref{tab:examples_qbcoref}). Across domains, the strategies behave
similarly, but we notice some differences in \textbf{ment-ent} and
\textbf{joint-ent}.  In \preco{}, those strategies tend to sample a mix of spans
that are and are not entity mentions (Section~\ref{ssec:distribution}).  In \qbcoref{}, they sample more entity
mentions. This could be due to more entity mentions present in a Quizbowl
question, which makes it more likely to sample something that should belong to
an entity cluster.

For other strategies, we notice some issues. As mentioned in Section~\ref{ssec:error}, \textbf{li-clust-ent} tends to
sample nested entity mentions, which may become redundant for annotators to
label. In fact, \avgfone{} for \textbf{li-clust-ent} tends to be lower if
document reading is constrained to one document. \textbf{Cond-ent} suffers from
redundant labeling because pronouns are repeatedly sampled and they tend to
link to the same entity cluster.

\begin{table*}[!t]
    \centering
    \small
    \renewcommand{\arraystretch}{1.6}
    \begin{tabular}{p{1.25cm}p{9.5cm}p{4cm}}
    \toprule
         Strategy & Sampled Spans & Comments\\
    \midrule
        \textbf{random} &
        Later, I got out of the back door secretly and gave the food to the old man, whose
        \entity{name I had discovered}{1}  was Taff. I had never seen anything else as lovely as the smile of
        satisfaction \entity{on}{2}  Taff's face when he ate the food. From then on, my visits to
        \entity{the old house had}{3} a purpose, and I enjoyed every minute of
        the rest of my stay. &
        \emph{Sampled spans are typically not entity mentions.} \\
        \textbf{random-ment} &
        When opening
        the door, his face was full of smiles and he hugged \entity{his two
        children and gave \entity{his
        wife}{2} a kiss}{1}. Afterwards, he walked with me to the car. We passed the tree. I
        was so curious that I asked  \entity{him}{3} about what I had
        \entity{seen}{4} earlier. &
        \emph{Diverse set of span types is sampled, including spans that
        are not entity mentions and ones that do link to entities.} \\
        \textbf{li-clust-ent} &
        Although \entity{he and \entity{his young men}{2}}{1} had taken no part
        in the killings, he knew that \entity{the white men}{3} would blame
        \entity{all of \entity{the Indians}{5}}{4}. &
        \emph{Many sampled spans are nested entity mentions.} \\
        \textbf{ment-ent} &
        This summer, Republicans have been  \entity{meeting}{1} ``behind closed
        doors'' on a Medicare proposal scheduled to be released \entity{later this month, only a few
        weeks before Congress votes}{2} on it, thereby avoiding independent analysis of the costs,
mobilization by opponents and other inconvenient aspects of a long national debate. Two years ago,
        the Republicans rang alarms about the \entity{Clinton}{3} plan's
        emphasis on \entity{managed
        care}{4} &
        \emph{Sampled spans are both entity mentions and non-entities. The spans
        are difficult for mention detection like ``meeting'' but may also be
        hard for clustering like ``Clinton''.} \\
        \textbf{clust-ent} &
        After that, \entity{Mary}{1}
        buys some school things, too. Here \entity{mother}{2} buys a lot of food, like bread,
        cakes, meat and fish. \entity{They}{3} get home very late. &
        \emph{Different types of entity mentions are sampled.} \\
        \textbf{cond-ent} &
        It is a chance to thank everyone who has contributed to
        shaping \entity{you}{1} during the high school years; it is a chance to appreciate all those
        who have been instrumental in \entity{your}{2} education. Take a moment to express gratitude
        to all those who have shared the experiences of \entity{your}{3} high
        school years. &
        \emph{More pronouns are sampled because
        they are obviously entity mentions and hard to cluster. However,
        repeated sampling of the same entity occurs.} \\
        \textbf{joint-ent} &
        \entity{This}{1} is an eternal regret handed down
        from generation to generation and \entity{you}{2} are only one of those who languish for
        (...) followers. \entity{Love}{3} is telephone, but it is difficult to seize
        \entity{the center time for dialing}{4}, and you will let the
        opportunity slip if your call is either too early or too late. &
        \emph{Many entity mentions are sampled but some are difficult for
        mention detector to detect.} \\
    \bottomrule
    \end{tabular}
    \caption{The example spans from \preco{} documents that are sampled with each active learning
    strategy.}
    \label{tab:examples_preco}
\end{table*}

\begin{table*}[!t]
    \centering
    \small
    \renewcommand{\arraystretch}{1.6}
    \begin{tabular}{p{1.25cm}p{9.5cm}p{4cm}}
    \toprule
         Strategy & Sampled Spans & Comments\\
    \midrule
        \textbf{random} &
        The discovery of a tube behind a \entity{fuse box alarms
        Linda, and the image of stock\entity{ings}{2} disturbs the main}{2} character due to his guilt
        over  \entity{an encounter with a woman and his son Biff in
        \entity{Boston}{4}}{3}. &
        \emph{Choice of sampled spans are very random and do not seem to improve
        learning coreference.} \\
        \textbf{random-ment} &
        The speaker of one of \entity{this author's works}{1} invites the
        reader to  \entity{take}{2} a little sun, a little honey, as commanded
        by \entity{Persephone's}{3} bees. &
        \emph{Diverse set of span types is sampled, including spans that
        are not entity mentions and ones that do link to entities.} \\
        \textbf{li-clust-ent} &
        For 10 points, name  \entity{this \entity{Moliere}{2} play about \entity{Argan who is constantly concerned
        with \entity{his}{4} health}{3}}{1}. &
        \emph{Many sampled spans are nested entity mentions.} \\
        \textbf{ment-ent} &
        He then sees \entity{Ignorance and Want}{1} emerge from \entity{a
        cloak}{2}. Earlier, he sees \entity{a door-knocker}{3}
        \entity{transform}{4} into
        \entity{a human figure, which drags a belt made of chains and
        locks}{5}. &
        \emph{Compared to \preco{}, more entity mentions are sampled but most
        sampled spans are still difficult to detect.} \\
        \textbf{clust-ent} &
        \entity{\entity{Its}{2} protagonist}{1} hires Croton to rescue a different character after listening to a giant - LRB -
        \ * - RRB - Christian named Urban \entity{discuss}{3} a meeting at
        Ostranium. &
        \emph{Compared to \preco{}, a few sampled spans are not entity mentions.} \\
        \textbf{cond-ent} &
        While \entity{this work}{1} acknowledges the
soundness of the arguments that use the example of the ancients,
        \entity{\entity{its}{3} author}{2} refuses to reply to
        \entity{them}{4}, adding that we are constructing no
        system here \entity{we}{5} are a historian, not a critic. &
        \emph{More pronouns are sampled because
        they are obviously entity mentions and hard to cluster. Unlike \preco{},
        repeated sampling occurs less often.} \\
        \textbf{joint-ent} &
        This man falls in love with \entity{the maid with \entity{lime colored
        panties}{2}}{1} and dates \entity{Luciana}{3}. &
        \emph{Compared to \preco{}, more entity mentions are sampled.} \\
    \bottomrule
    \end{tabular}
    \caption{The example spans from \qbcoref{} documents that are sampled with each active learning
    strategy.}
    \label{tab:examples_qbcoref}
\end{table*}

\end{document}